\documentclass[10pt,twocolumn,letterpaper]{article}

\usepackage[pagenumbers]{cvpr} 

\definecolor{cvprblue}{rgb}{0.21,0.49,0.74}
\definecolor{myblue}{HTML}{6C8EBF}
\definecolor{myred}{HTML}{B85450}
\usepackage{lipsum}
\usepackage{booktabs,tabularx}
\usepackage{cuted}   
\usepackage[table, dvipsnames]{xcolor}
\usepackage{adjustbox}
\usepackage{pifont}         
\usepackage{multicol}
\usepackage{multirow}
\usepackage[ruled,vlined]{algorithm2e}
\usepackage{tcolorbox}
\usepackage{verbatim}
\usepackage{tabularx}
\tcbuselibrary{listings, breakable}
\usepackage{graphicx}
\usepackage{mathrsfs}
\usepackage{float}

\definecolor{green_t}{rgb}{0.267,0.416,0.224}
\definecolor{green_a}{rgb}{0.227, 0.714, 0.227}
\definecolor{road}{rgb}{0.875, 0.851, 0.812}
\definecolor{water}{rgb}{0.149, 0.322, 0.671}
\definecolor{building}{rgb}{0.796, 0.416, 0.345}
\definecolor{barren}{rgb}{0.788, 0.635, 0.459}

\newcommand{\cmark}{\ding{51}}
\newcommand{\xmark}{\ding{55}}
\usepackage[pagebackref,breaklinks,colorlinks,allcolors=cvprblue]{hyperref}

\def\paperID{{35637}} 
\def\confName{CVPR}
\def\confYear{2026}

\title{Enabling Training-Free Text-Based Remote Sensing Segmentation}

\author{
Jose Sosa, Danila Rukhovich, Anis Kacem, Djamila Aouada \\
SnT, University of Luxembourg \\
{\tt\small \{jose.sosa,danila.rukhovich,anis.kacem,djamila.aouada\}@uni.lu}
}

\begin{document}

\maketitle

\begin{strip}
\centering
\includegraphics[width=\linewidth]{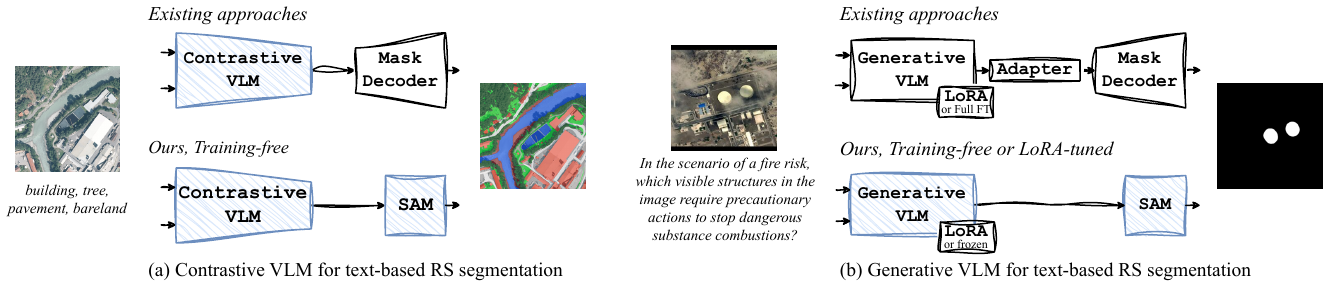}
\captionof{figure}{Existing methods~\cite{li2025segearth, li2025segearth-r1, shabbir2025geopixel} rely on additional trainable mask decoders and adapters. We propose a training-free methodology that combines VLMs and SAM without introducing new trainable components. Additionally, with LoRA fine-tuning, our method achieves state-of-the-art performance on reasoning segmentation. \textcolor{myblue}{Blue} is for frozen components.}
\label{fig:teaser}
\end{strip}

\begin{abstract}

\vspace{-0.5em}
Recent advances in Vision Language Models (VLMs) and Vision Foundation Models (VFMs) have opened new opportunities for zero-shot text-guided segmentation of remote sensing imagery. However, most existing approaches still rely on additional trainable components, limiting their generalisation and practical applicability. In this work, we investigate to what extent text-based remote sensing segmentation can be achieved without additional training, by relying solely on existing foundation models. We propose a simple yet effective approach that integrates contrastive and generative VLMs with the Segment Anything Model (SAM), enabling a fully training-free or lightweight LoRA-tuned pipeline. Our contrastive approach employs CLIP as mask selector for SAM’s grid-based proposals, achieving state-of-the-art open-vocabulary semantic segmentation (OVSS) in a completely zero-shot setting. In parallel, our generative approach enables reasoning and referring segmentation by generating click prompts for SAM using GPT-5 in a zero-shot setting and a LoRA-tuned Qwen-VL model, with the latter yielding the best results. Extensive experiments across 19 remote sensing benchmarks, including open-vocabulary, referring, and reasoning-based tasks, demonstrate the strong capabilities of our approach. Code will be released \href{https://github.com/josesosajs/trainfree-rs-segmentation}{here}.

\end{abstract}

\section{Introduction}
Remote sensing imagery has become a cornerstone of earth observation, supporting critical applications such as land-cover mapping, environmental monitoring, and disaster response~\cite{jakubik2023prithvi, szwarcman2024prithvi2, sosa2024effective, sosa2025multimae}. Recent advances in deep learning, in particular for pixel-level segmentation have highly improved the accuracy and scalability of such analyses~\cite{zhu2025skysense-o, dong2025diffris}. However, most methods still follow supervised setups, depending on large-scale, domain-specific annotated datasets for training. The costly and inconsistent process of collecting dense pixel-level annotations continues to limit progress, particularly for fine-grained or rapidly evolving geospatial categories.

Recently, Vision Language Models (VLMs)~\cite{radford2021clip, bai2025qwen2.5vl, achiam2023gpt-4} and Vision Foundation Models (VFMs)~\cite{ravi2024sam} have shown impressive zero-shot capabilities on natural images, achieving text-based segmentation without additional supervision. These models offer an appealing direction for remote sensing, where various successful approaches have emerged~\cite{li2025segearth, li2025segearth-r1, shabbir2025geopixel, zhou2024geoground, lai2024lisa, zhang2023next, ren2024pixellm, yang2022lavt, zhou2022extract, bai2024video-lisa}. Nevertheless, most methods in this domain still rely on additional trainable adapters~\cite{li2025segearth-r1, li2025mask, zhou2022extract}, lightweight heads~\cite{li2025segearth, zhang2023next}, or token-level bridges~\cite{lai2024lisa, bai2024video-lisa, ren2024pixellm} to link visual and textual modalities.

Our work is based on the premise that relying exclusively on pretrained foundation models enables a training-free approach for text-based remote sensing segmentation. This raises a central question: \textit{To what extent can such segmentation be achieved solely through pretrained foundation models, without introducing any additional trainable components?} To explore this, our approach builds on two key elements: VLMs and VFMs. VLMs provide the multi-modal link between textual queries and visual content, while VFMs (such as SAM~\cite{ravi2024sam}), offer a generic mechanism for mask generation. We investigate strategies to integrate these components without introducing additional trainable parameters. 

Specifically, we propose two complementary pipelines, the first uses \textit{contrastive VLMs}, like CLIP~\cite{radford2021clip}, as semantic selectors over SAM’s category-agnostic mask proposals for OVSS. The second employs \textit{generative VLMs}, such as GPT-5~\cite{OpenAI2025GPT5} and Qwen-VL~\cite{bai2025qwen2.5vl}, as SAM prompters with spatial clicks for referring and reasoning-based segmentation scenarios. While the first approach operates in a fully zero-shot manner, the second can be applied either zero-shot or with lightweight LoRA fine-tuning~\cite{hu2022lora}.~\autoref{fig:teaser} provides a visual comparison of our proposed approach, and contrasts it with recent text-based remote sensing segmentation methods. In summary, our contributions are as follows:

\begin{itemize}
\item We investigate the extent to which text-based remote sensing segmentation can be accomplished by using only existing VLMs and SAM, without introducing additional trainable components.
\item We propose two complementary approaches for combining VLMs with SAM: (i) using a contrastive VLM to select masks from SAM’s grid-based proposals, and (ii) using a generative VLM to generate click prompts for SAM-based segmentation.
\item We demonstrate that contrastive VLM-based pipeline enables fully training-free segmentation, achieving state-of-the-art performance in OVSS of remote sensing imagery.
\item We further show that minimal LoRA fine-tuning of the generative VLM-based approach, with SAM kept frozen, yields state-of-the-art results on reasoning and referring segmentation with remote sensing imagery.
\end{itemize}


\section{Related work}
\label{sec-rws}
\subsection{VLMs for Text-Based Segmentation}
Text-based image segmentation aims to segment regions within an image based on natural language descriptions. Recent advances~\cite{lai2024lisa, ren2024pixellm, zhang2023next} on multi-modal datasets and pretraining strategies for VLMs have revolutionised this field. Consequently, text-based segmentation is increasingly dominated by contrastive and generative VLM-based approaches. These models can localise complex language-guided targets without requiring extensive task-specific supervision.

\vspace{0.5em}
\noindent\textbf{Contrastive VLMs}~\cite{radford2021clip,luo2023segclip} are trained to align image and text representations through contrastive learning on paired data. Starting from CLIP~\cite{radford2021clip}, they have shown remarkable zero-shot performance on image classification, which naturally extends to semantic segmentation~\cite{luddecke2022image}. These approaches are usually categorised based on the amount of needed supervision. \textit{Training-free} methods attempt to exploit the inherent localisation capabilities of CLIP with minimal modifications. For instance, MaskCLIP~\cite{zhou2022extract} proposes to extract the value embedding of the last self-attention block of CLIP’s vision encoder for dense prediction tasks. Following this work, other studies~\cite{lan2024clearclip} generalise the query-key attention to a self-self attention mechanism. This includes, the value-value attention in CLIPSurgery~\cite{li2025closer}, the query-query and key-key attention in SCLIP~\cite{wang2024sclip}, and generalised self-self attention combination in GEM~\cite{bousselham2024grounding}. Another stream of work~\cite{barsellotti2024training, kang2024defense, shao2024explore, sun2024clip, wang2025diffusion} adopts a two-stage method. The first stage generates category-agnostic mask proposals, while the second stage classifies them. \textit{Trainable} methods allow models to be trained on some base classes in a supervised or weakly supervised manner. Typically, some works ~\cite{ghiasi2022scaling, luo2023segclip, mukhoti2023open, ranasinghe2023perceptual} train a localisation-aware CLIP for dense predictions. Others ~\cite{cho2024cat, ding2022decoupling, li2022language, liu2024open, xu2023san} instead fine-tune a subset of CLIP’s pre-trained parameters or add a few trainable ones to adapt it for dense prediction on base classes. Finally, with additional adapters~\cite{li2025mask, barsellotti2025talking, wang2024sam, jiao2023learning, lan2024proxyclip, yang2024tuningfree,kang2024defenselazyvisualgrounding} CLIP can be connected with other foundation models (SAM~\cite{ravi2024sam} or DINO~\cite{caron2021emerging}) to enhance the localisation ability.

\vspace{0.5em}
\noindent\textbf{Generative VLMs}~\cite{bai2025qwen2.5vl, xu2025qwen3-omni, achiam2023gpt-4} are trained to model the joint or conditional distribution of images and text via auto-regressive generation. These models allow for complex reasoning between image and language modalities. However, none of them directly support image segmentation, so need to be extended with extra components. LISA~\cite{lai2024lisa} establishes the paradigm by introducing a \texttt{<SEG>} token to connect LLMs with segmentation decoders. Finetuning VLMs with these novel tokens is further explored in PixelLM~\cite{ren2024pixellm} and more recent works~\cite{zhang2024omg, rasheed2024glamm, zhang2024psalm, wang2024llm-seg, zhang2023next, yuan2025sa2va, bai2024video-lisa, wang2025x-sam}. SAM4MLLM~\cite{chen2024sam4mllm} and SegAgent~\cite{zhu2025segagent} propose the solution without adding novel trainable tokens, namely to use SAM~\cite{ravi2024sam} as a tool, conditioned by clicks or boxes, produced by VLM in form of text. Another possible tool for image generation for VLMs can be diffusion models, \eg Qwen-Image~\cite{wu2025qwen-image} for Qwen3~\cite{xu2025qwen3-omni} or GPT-Image-1~\cite{openai_image_generation_2025} for GPT-5~\cite{OpenAI2025GPT5}. This technically gives VLMs the ability to solve segmentation, since segmentation mask can be imagined as just an image. 

Overall, despite the proven capabilities of aforementioned contrastive and generative VLM-based methods on natural images, none of them originally attempted text-based remote sensing segmentation.

\begin{figure}[h]
    \centering
    \includegraphics[width=0.935\linewidth]{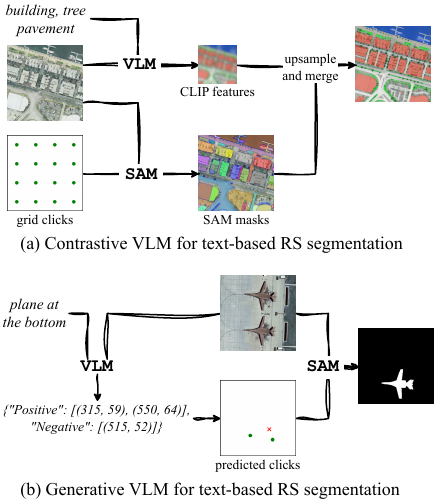}
    \caption{Inference schemes of our segmentation approaches with (a) contrastive and (b) generative VLMs.}
    \label{fig:main-method}
\end{figure}

\subsection{Text-Based Remote Sensing Segmentation}
Early studies on text-based remote sensing segmentation primarily rely on masked language models such as BERT~\cite{devlin2019bert} to encode textual inputs. These models are typically followed by a task-specific conditional convolutional decoder~\cite{yuan2023rrsis, liu2024rotated, lei2024fianet, chen2025rsrefseg, li2025sbanet, li2025aeroreformer, zhang2025btdnet, zhu2025skysense-o} or diffusion-based architectures such as DiffRIS~\cite{dong2025diffris}. Despite notable progress, these approaches remain heavily supervised and rely on dedicated model designs for specific datasets or prompt types. Recently, SegEarth-OV~\cite{li2025segearth} marks a shift towards reducing training dependency by introducing a nearly training-free framework. Their method incorporates a frozen CLIP model for image–text alignment, while training only an unsupervised mask decoder. They further explore remote sensing adaptations of CLIP, such as GeoRSCLIP~\cite{zhang2024georsclip} and RemoteCLIP~\cite{liu2024remoteclip}, which improve visual grounding in geospatial imagery. 

A contemporary line of research transitions to generative VLMs to handle more complex referring and reasoning prompts. For example, GeoGround~\cite{zhou2024geoground} reformulates segmentation as per-tile binary prediction using a VLM backbone without an explicit decoder. An extended version incorporates multiple auxiliary encoders and a dedicated mask decoder for finer spatial resolution. SegEarth-R1~\cite{li2025segearth-r1} further advances this line of work by introducing a reasoning-focused remote sensing benchmark. It also proposes a simplified yet trainable mask encoder that is conditioned jointly on the vision and language embeddings.

Typically, existing methods achieve language-conditioned segmentation by adding trainable components on top of VLMs. These components may include mask decoders, adapters, or token-level bridges. In contrast, our work shows that combining only a VLM with SAM is sufficient to achieve state-of-the-art results across open-vocabulary, referring, and reasoning segmentation tasks in remote sensing, without any additional trainable modules.

\section{Methodology}
\label{sec-method}
In this work, we address the task of \textit{text-based remote sensing image segmentation}. Given an image $I$ and a textual query $t$, the goal is to produce a segmentation mask $M$ that corresponds to the region in the image described by the text. Our objective is to design a solution that requires minimal training, and ideally operates in a fully zero-shot manner, \ie, without any task-specific training.

To process textual instructions and input images, we adopt VLMs that are already pre-trained on large-scale image–text pairs in a self-supervised manner. These models have demonstrated strong zero-shot generalisation across a wide range of vision–language tasks. However, existing VLMs do not inherently generate segmentation masks, limiting their direct applicability to segmentation tasks. On the other hand, VFMs such as SAM~\cite{ravi2024sam} have shown remarkable performance across various segmentation tasks, including semantic, instance, and panoptic segmentation. As illustrated on~\autoref{fig:main-method}, we propose to combine these powerful pre-trained components, a VLM for understanding text and images, and SAM for mask generation. This combination does not require any additional trainable modules, enabling a fully training-free paradigm for text-based remote sensing segmentation.

VLMs can be broadly categorised into two types. The first type is \textit{contrastive} models, such as CLIP~\cite{radford2021clip}, which are trained to align images and text in a shared embedding space. The second type is \textit{generative} models, such as Qwen-VL~\cite{bai2025qwen2.5vl}, which are trained to autoregressively generate text conditioned on visual inputs. In the following sections, we describe how we use both types of VLMs in combination with SAM to solve the task of text-based remote sensing segmentation.

\subsection{Contrastive VLMs as SAM Mask Selectors}

\paragraph{Pipeline.} Let $\mathscr{F}$ denote a contrastive VLM and $\mathscr{S}$ the SAM model. Given an input image $I$ and a textual prompt $t$, the contrastive VLM processes $I$ and $t$ and computes a per-pixel foreground probability map $p(x,y)$ for $t$. In parallel, SAM produces a set of $K$ category-agnostic mask proposals $\{M_k\}_{k=1}^{K}$ given $I$ and a set of clicks $\mathcal{C}$ in form of a regular 2D grid.

For each SAM mask $M_k$, we determine whether it corresponds to the target object by counting the proportion of pixels with $p(x,y) > 0.5$ with the following indicator function, 
\[
\delta_k =
\begin{cases}
1, & \text{if } \frac{1}{|M_k|} \sum_{(x,y) \in M_k} \mathbf{1}[p(x,y) > 0.5] > 0.5  \ , \\
0, & \text{otherwise}.
\end{cases}
\]
The final prediction is obtained by merging all relevant masks as follows, 
\[
M = \bigcup_{k=1}^{K} \{ M_k \mid \delta_k = 1 \}  \ .
\]

\vspace{0.5em}
\noindent\textbf{Extension to multi-class segmentation.} Given $m$ text prompts $\{t_i\}_{i=1}^{m}$, the VLM predicts per-pixel probabilities $p_i(x,y)$ for each class. Each pixel is assigned to the class with the highest probability as follows, 
\[
\text{class}(x,y) = \arg\max_{i \in \{1,\dots,m\}} p_i(x,y) \ .
\]

To mitigate CLIP’s global bias, we apply the debiasing technique from~\cite{li2025segearth}, which subtracts a scaled \texttt{<CLS>} token from all patch tokens. Each SAM mask $M_k$ is then assigned to the class that dominates within its area, 
\[
\ell_k = \arg\max_{i \in \{1,\dots,m\}} \left| \{ (x,y) \in M_k \mid \text{class}(x,y) = i \} \right| \ .
\]
The final segmentation for class $i$ is expressed as:
\[
M^i = \bigcup_{k=1}^{K} \{ M_k \mid \ell_k = i \} \ .
\]
Note that masks or pixels not assigned to any class remain background.

\subsection{Generative VLMs as SAM prompters}
\paragraph{Pipeline.} To avoid introducing any trainable components between the VLM and SAM, the only way to condition SAM with language is by expressing its prompts in text. SAM supports two types of lightweight prompts (\textit{clicks} and \textit{bounding boxes}) that can be easily described with text. For consistency with our contrastive-VLM approach, we focus only on click-based prompting.

Let the generative VLM be denoted as $\mathscr{F}$. Given an image $I$ and a textual instruction $t$, the VLM outputs a set of clicks $\mathcal{C} = \{c_i\}_{i=1}^n$ that indicate the target region
\[
    \mathcal{C} = \mathscr{F}(I, t) \ .
\]
These clicks serve as prompts for SAM ($\mathscr{S}$), which generates the final segmentation mask $M$:
\[
    M = \mathscr{S}(I, \text{prompt} = \mathcal{C}) \ .
\]
Each click is labeled as either \textit{positive} or \textit{negative}, indicating whether the corresponding location should be included or excluded from the segmentation mask. In textual form, the click set is represented as:
\[
\begin{aligned}
\{\text{``Positive''} &:[(x_{+,1},y_{+,1}),(x_{+,2},y_{+,2}),\ldots], \\
\text{``Negative''} &:[(x_{-,1},y_{-,1}),(x_{-,2},y_{-,2}),\ldots]\} \ .
\end{aligned}
\]

If no negative clicks are present, SAM generates the mask using only the positive set.

\vspace{0.5em}
\noindent\textbf{Training.} Preliminary experiments show that this pipeline already achieves reasonable segmentation quality. This is observed in a fully zero-shot manner when $\mathscr{F}$ is a large proprietary VLM. However, to further improve performance and generalisation, we propose to fine-tune a smaller open-source generative VLM for click generation, while keeping SAM completely frozen.

To train the VLM, we simply concatenate the image $I$, textual instruction $t$, and the click sequence $\mathcal{C}$ in text form into a single token sequence. The model is then optimised using standard next-token prediction (cross-entropy loss). However, the missing component is the supervision signal, since existing segmentation datasets provide masks $M$ but not click annotations. We address this by automatically converting masks into click sequences, as described below.

\vspace{0.5em}
\noindent\textbf{Training clicks generation.} Existing text-based image segmentation datasets usually provide ground-truth annotations in the form of per-pixel masks. Our goal is to convert these masks $M$ into a sequence of clicks $\mathcal{C}$ without human involvement. To do so, we adopt an iterative strategy inspired by interactive segmentation~\cite{sofiiuk2022reviving, antonov2024rclicks}. 

Starting from an image and its corresponding ground-truth mask, SAM is prompted with an initial positive click inside the target object to produce a mask. This prediction is then compared with the ground-truth mask to identify under-segmented and over-segmented regions. Additional clicks are placed in these regions, positive clicks in missing areas and negative clicks in unwanted regions. Then, SAM is prompted again to update the mask. This process is repeated until a stopping condition is met (\eg, achieving a target IoU or reaching a maximum number of clicks). The resulting synthetic click sequences $\mathcal{C}$ are then used to finetune the generative VLM for click generation. More details about this process are given in Appendix.



\subsection{Application to Text-Based Remote Sensing Segmentation}

In text-based remote sensing segmentation, text prompts vary significantly in complexity. Existing settings can be grouped into three categories: (i) \textbf{OVSS:} each class is described using a short phrase or a couple of keywords (\eg, \textit{road}, \textit{industrial area}). (ii) \textbf{Referring segmentation:} each prompt is a full sentence that describes a specific region or object within the image (\eg, \textit{The vehicle on the upper right}). (iii) \textbf{Reasoning segmentation:} the prompt requires multi-step reasoning or implicit understanding, without explicitly naming the target region (\eg, \textit{Which part of the infrastructure is best for rapid patient transport by emergency services?}).

Our contrastive and generative VLM-based pipelines naturally align with these three levels of complexity. Contrastive VLMs perform well with short, unambiguous prompts, making them suitable for OVSS. However, their capability degrades when prompts become longer, descriptive, or require contextual reasoning. Alternatively, generative models are better at understanding complex linguistic instructions and grounding them spatially through click prompts. Therefore, we employ the generative VLM approach for referring and reasoning-based segmentation. 

This design choice raises an important question: \textit{why not use the generative approach for all three tasks?} The limitation lies in the nature of how generative VLMs interact with SAM. These models typically produce only a small set of clicks, resulting in a single (or very few) connected mask. This is adequate for referring and reasoning segmentation tasks, where usually only one instance is expected. However, in OVSS, many semantic categories such as \textit{forest}, \textit{water}, or \textit{urban area} might consist of multiple spatially disconnected regions. A single SAM prompt, even with various positive and negative clicks, often fails to capture all relevant areas when SAM is kept frozen. Consequently, OVSS requires combining multiple SAM-generated masks, which our contrastive VLM-based mask selection approach naturally enables. In summary, contrastive VLMs are preferable for OVSS tasks, while generative VLMs are more appropriated for referring and reasoning-based segmentation.

\begin{table*}
\centering
\begin{tabular}{llcccccccc|c}
\toprule
& Method & OEM & LoveDA & iSAID & Potsdam & Vaihingen & UAVid & UDD5 & VDD & Avg. \\
\midrule
\multicolumn{3}{l}{\textit{Trained on remote sensing data}} \\
& SegEarth-OV~\cite{li2025segearth} & 40.3 & 36.9 & 21.7 & 48.5 & 40.0 & 42.5 & 50.6 & 45.3 & 39.2 \\
& \textcolor{gray}{Oracle} & \textcolor{gray}{64.4} & \textcolor{gray}{50.0} & \textcolor{gray}{36.2} & \textcolor{gray}{74.3} & \textcolor{gray}{61.2} & \textcolor{gray}{59.7} & \textcolor{gray}{56.5} & \textcolor{gray}{62.9} & \textcolor{gray}{58.2} \\
\midrule
\multicolumn{3}{l}{\textit{Zero-shot methods}} \\
& CLIP~\cite{radford2021clip} & 12.0 & 12.4 & 7.5 & 15.6 & 10.8 & 10.9 & 9.5 & 14.2 & 11.4 \\
& MaskCLIP~\cite{zhou2022extract} & 25.1 & 27.8 & 14.5 & 33.9 & 29.9 & 28.6 & 32.4 & 32.9 & 27.2 \\
& SCLIP~\cite{wang2024sclip} & 29.3 & 30.4 & 16.1 & 39.6 & 35.9 & 31.4 & 38.7 & 37.9 & 31.1 \\
& GEM~\cite{bousselham2024grounding} & 33.9 & 31.6 & 17.7 & 39.1 & 36.4 & 33.4 & 41.2 & 39.5 & 32.3 \\
& ClearCLIP~\cite{lan2024clearclip} & 31.0 & 32.4 & 18.2 & 42.0 & 36.2 & 36.2 & 41.8 & 39.3 & 33.4 \\
& \textbf{Ours} & \textbf{34.2}  &\textbf{38.2} & \textbf{21.9} & \textbf{50.2} & \textbf{40.6} & \textbf{44.3} & \textbf{53.8} & \textbf{46.8} & \textbf{41.3} \\
\bottomrule
\end{tabular}
\caption{Results of our contrastive VLM-based approach for text-based remote sensing segmentation on OVSS task. We evaluate 8 remote sensing multi-class datasets. \textit{Avg.} is for average across all datasets. \textit{Oracle} represents the upper bound, achieved by a fully supervised model~\cite{xie2021segformer}.}
\label{tab:comparison-ov-sem-seg-mc}
\end{table*}

\begin{table*}
\centering
\resizebox{\textwidth}{!}{
\begin{tabular}{llccccccccc|c}
\toprule
& \multirow{2}{*}{Method} & \multicolumn{4}{c}{Building extraction} & \multicolumn{4}{c}{Road Extraction} & Flood Detection & \multirow{2}{*}{Avg.} \\
& & WHU-A & WHU-S & Inria & xBD-pre & CHN6 & DG & MA & SpaceNet & WBS-SI & \\
\midrule
\multicolumn{3}{l}{\textit{Trained on remote sensing data}} \\
& SegEarth-OV~\cite{li2025segearth} & 49.2 & 28.4 & 44.6 & 37.0 & 35.4 & 17.8 & 11.5 & 23.8 & 60.2 & 34.2 \\
\midrule
\multicolumn{3}{l}{\textit{Zero-shot methods}} \\
& CLIP~\cite{radford2021clip} & 17.7 & 3.5 & 19.6 & 16.0 & 7.7 & 3.9 & 4.9  & 7.1 & 18.6 & 11.0 \\
& MaskCLIP~\cite{zhou2022extract} & 29.8 & 14.0 & 33.4 & 29.2 & 28.1 & 13.2 & 10.6 & 20.8 & 39.8 & 24.3 \\
& SCLIP~\cite{wang2024sclip} & 33.4 & 21.0 & 34.9 & 25.9 & 21.1 & 7.0  & 7.4 & 14.9 & 32.1 & 22.0 \\
& GEM~\cite{bousselham2024grounding} & 24.4 & 13.6 & 28.5 & 20.8 & 13.4 & 4.7 & 5.1 & 11.9 & 39.5 & 18.0 \\
& ClearCLIP~\cite{lan2024clearclip} & 36.6 & 20.8 & 39.0 & 30.1 & 25.5 & 5.7 & 6.4 & 16.3 & 44.9 & 25.0 \\
& \textbf{Ours} & \textbf{58.8} & \textbf{26.1} & \textbf{48.0} & \textbf{34.4} & \textbf{36.4} & \textbf{15.9} & \textbf{12.2} & \textbf{26.1} & \textbf{58.3} & \textbf{35.1} \\
\bottomrule
\end{tabular}
}
\caption{Results of our contrastive VLM-based approach for text-based remote sensing segmentation on OVSS task. We evaluate 9 remote sensing single-class datasets across building extraction, road extraction, and flood detection. \textit{Avg.} is for average across all datasets.}
\label{tab:dataset_sem-seg-2c}
\end{table*}

\section{Experiments}
\label{sec-exp}

\subsection{Datasets and Implementation Details}

\paragraph{OVSS.} Following prior works~\cite{li2025segearth}, we evaluate our approach on 17 widely used datasets for multi-class and single-class remote sensing semantic segmentation. For multi-class standard semantic segmentation, we use 5 datasets depicting satellite images including OpenEarthMap~\cite{xia2023openearthmap}, LoveDA~\cite{wang2021loveda}, iSAID~\cite{waqas2019isaid}, Potsdam, and Vaihingen~\cite{ISPRS_UrbanSemLab}. We also consider 3 additional datasets containing UAV images, UAVid~\cite{lyu2020uavid}, UDD5~\cite{chen2018large}, and VDD~\cite{cai2025vdd}. In the context of single-class semantic segmentation, the datasets depict two classes: the foreground class, corresponding to building, road or water respectively, and the background class. We employ 4 datasets for building extraction, WHUAerial~\cite{ji2018fully}, WHUSat.II~\cite{ji2018fully}, Inria~\cite{maggiori2017can}, and xBD~\cite{gupta2019xbd}; 4 for road extraction, CHN6-CUG~\cite{zhu2021global}, DeepGlobe~\cite{DeepGlobe2018}, Massachusetts~\cite{mnih2013machine}, and SpaceNet~\cite{van2018spacenet}; and 1 for flood detection, WBS-SI~\cite{Kaggle_WaterBodySegmentation}. More details about datasets could be found in Appendix.

\vspace{0.5em}
\noindent\textbf{Referring and reasoning segmentation.} For referring segmentation, we use the widely adopted RRSIS-D dataset~\cite{liu2024rotated}. RRSIS-D includes 17,402 image–description-mask triplets, divided into 12,181 for training, 1,740 for validation, and 3,481 for testing. For the reasoning segmentation task, we adopt the large-scale EarthReason benchmark~\cite{li2025segearth-r1}. This dataset contains 5,434 images, each associated with an average of six questions and corresponding masks. The dataset is split into 2,371, 1,135, and 1,928 images for the training, validation, and test sets, respectively. For both datasets, we report metrics on validation and test sets. Train split is used only for click generation and VLM fine-tuning. 

\vspace{0.5em}
\noindent\textbf{Implementation details.} For our contrastive VLM-based approach, we user CLIP-base~\cite{radford2021clip} for image and text encoding, and SAM-L~\cite{ravi2024sam} as the mask generator. We sample an uniform grid of $29 \times 29$ positive click points for the main experiments. Following ~\cite{li2025segearth}, all images are resized to $448 \times 448$ pixels for CLIP while SAM operates on images at their original resolution. Performance is measured with mIoU for multi-class and foreground IoU for single-class datasets. For the generative VLM-based approach, in the zero-shot setting, we utilise the GPT-Image-1 API~\cite{openai_image_generation_2025} and GPT-5~\cite{OpenAI2025GPT5} for image and clicks generation, respectively. For the fine-tuning setting, we adopt Qwen3-VL-2B~\cite{wu2025qwen-image} as the backbone model. Training is conducted on four A100 GPUs for 3 epochs, using batch size of 64, LoRA (rank 32), the AdamW optimiser with a learning rate of \mbox{2e-4} and a cosine learning rate scheduler. We employ a total of 6 clicks during training. For inference, the same SAM as in the contrastive setup is used. Performance is reported using mIoU for both referring and reasoning tasks; for the reasoning task, the final mask is obtained via average voting over six predictions per image.

\subsection{Comparison with Prior Work}
\vspace{0.5em}
\noindent\textbf{Enabling training-free OVSS.} We evaluate our contrastive VLM-based approach on multi-class semantic segmentation benchmarks. We compare it with zero-shot natural image baselines and SegEarth-OV~\cite{li2025segearth}, which is the closest prior work toward training-free OVSS for remote sensing. As shown in \autoref{tab:comparison-ov-sem-seg-mc}, our method achieves state-of-the-art zero-shot performance across datasets, demonstrating strong generalisation without task-specific training. 

Compared with analogous zero-shot baselines such as CLIP, our model shows a substantial performance gain, highlighting the effectiveness of combining CLIP with a frozen SAM. Against SegEarth-OV, our approach outperforms on 7 of 8 datasets, with the largest gains on UAV imagery (UAVid, UDD5, VDD). Similar to~\cite{li2025segearth}, it struggles on iSAID due to fine-grained categories, and on OpenEarthMap minor class confusions slightly reduce performance. Notably, unlike SegEarth-OV, which requires training auxiliary components on remote sensing data (SimFeatUp~\cite{fu2024featup}), our approach is entirely training-free. 

On 9 single-class extraction datasets (\autoref{tab:dataset_sem-seg-2c}), our approach achieves state-of-the-art results among zero-shot baselines and even surpasses~\cite{li2025segearth} on 5. Compared directly to~\cite{li2025segearth}, our method ranks first on half of the building extraction datasets and second on the rest. For road extraction, it outperforms~\cite{li2025segearth} on 3 datasets, though with smaller margins, likely due to the challenge of zero-shot localisation of thin, complex structures.

\begin{figure}[t]
\centering \footnotesize \setlength{\tabcolsep}{2pt}
\begin{tabular}{ccc}
Input image & Predicted mask & GT mask \\
\includegraphics[width=0.315\linewidth]{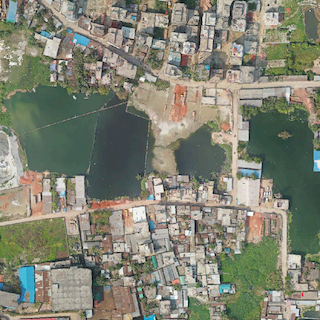} &
\includegraphics[width=0.315\linewidth]{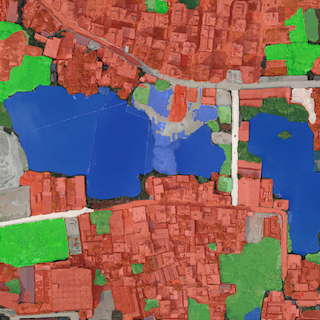} &
\includegraphics[width=0.315\linewidth]{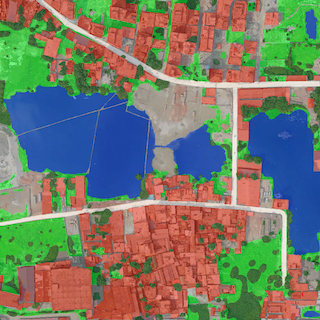}
\end{tabular}
\vspace{-9pt}
\begin{tcolorbox}[colback=gray!20, colframe=black!0, top=0.4pt, bottom=0.4pt]
\scriptsize
Prompt: \textit{\textcolor{gray}{Background}, Bareland, Rangeland, Developed Space, \textcolor{white}{Road}, \textcolor{green_t}{Tree}, \textcolor{water}{Water}, \textcolor{green_a}{Agriculture Land}, and \textcolor{building}{Building}.}
\end{tcolorbox}
\vspace{-3pt}
\begin{tabular}{ccc}
\includegraphics[width=0.315\linewidth]{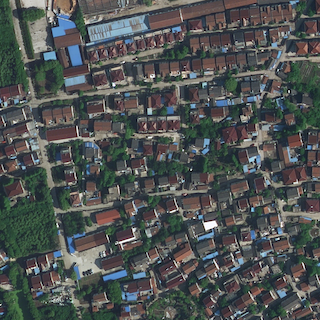} &
\includegraphics[width=0.315\linewidth]{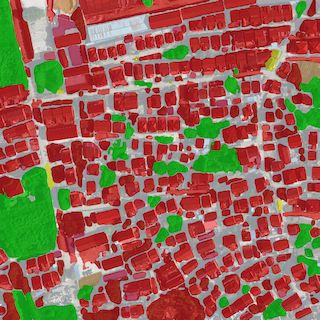} &
\includegraphics[width=0.315\linewidth]{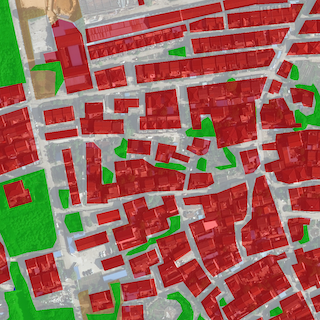}
\end{tabular}
\vspace{-9pt}
\begin{tcolorbox}[colback=gray!20, colframe=black!0, top=0.4pt, bottom=0.4pt]
\scriptsize
Prompt: \textit{Background, \textcolor{red}{Building}, \textcolor{gray}{Road}, Water, \textcolor{barren}{Barren}, \textcolor{green_a}{Forest}, and Agriculture.\phantom{pppfff}} 
\end{tcolorbox}
\vspace{-3pt}
\begin{tabular}{ccc}
\includegraphics[width=0.315\linewidth]{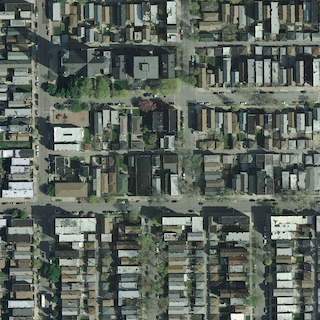} &
\includegraphics[width=0.315\linewidth]{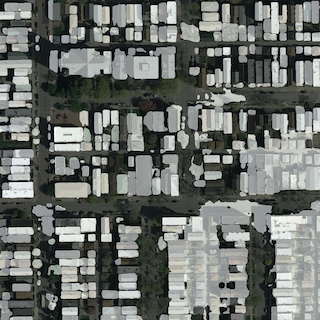} &
\includegraphics[width=0.315\linewidth]{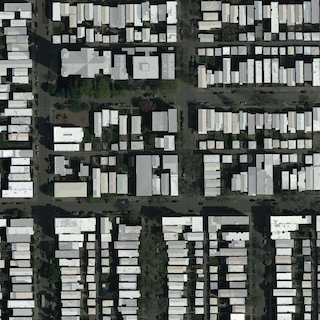}
\end{tabular}
\vspace{-9pt}
\begin{tcolorbox}[colback=gray!20, colframe=black!0, top=0.4pt, bottom=0.4pt]
\scriptsize
Prompt: \textit{Background and \textcolor{white}{Building}.}
\end{tcolorbox}   
\caption{Qualitative results of the training-free contrastive VLM pipeline on multi-class (first and second rows) and single-class (third row) OVSS tasks using remote sensing datasets.}
\label{fig:ovss-main}
\end{figure}

\begin{figure}[h]
\centering \footnotesize \setlength{\tabcolsep}{2pt}
\begin{tabular}{ccc}
Predicted clicks & Predicted mask & GT mask \\
\includegraphics[width=0.315\linewidth]{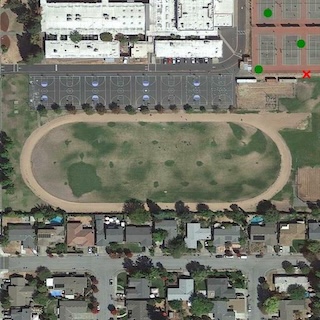} &
\includegraphics[width=0.315\linewidth]{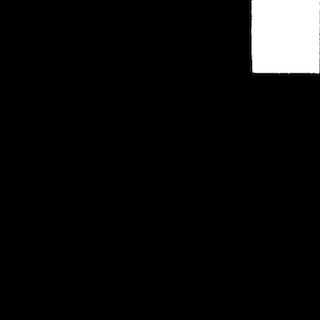} &
\includegraphics[width=0.315\linewidth]{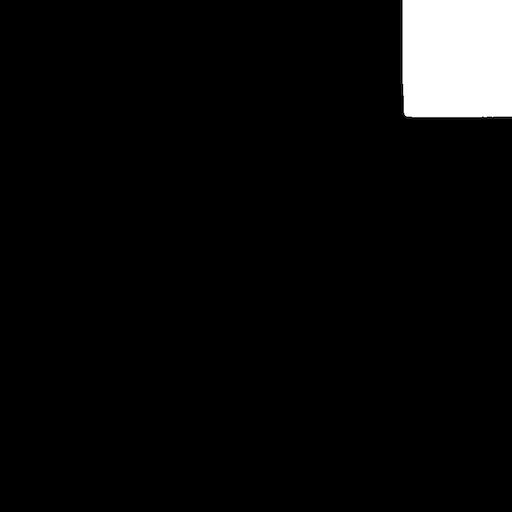}
\end{tabular}
\vspace{-9pt}
\begin{tcolorbox}[colback=gray!20, colframe=black!0, top=0.4pt, bottom=0.4pt]
\scriptsize
    Prompt: \textit{Should you wish to improve your overhead serve and join a doubles match, what specific location in the sports complex would you select?}
\end{tcolorbox}
\vspace{-3pt}
\begin{tabular}{ccc}
\includegraphics[width=0.315\linewidth]{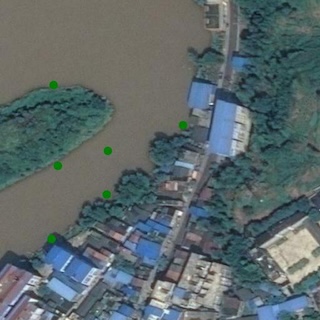} &
\includegraphics[width=0.315\linewidth]{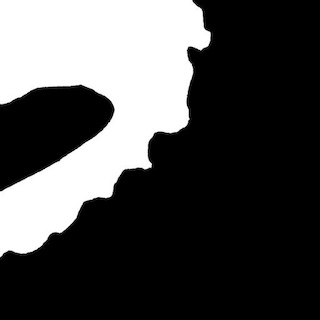} &
\includegraphics[width=0.315\linewidth]{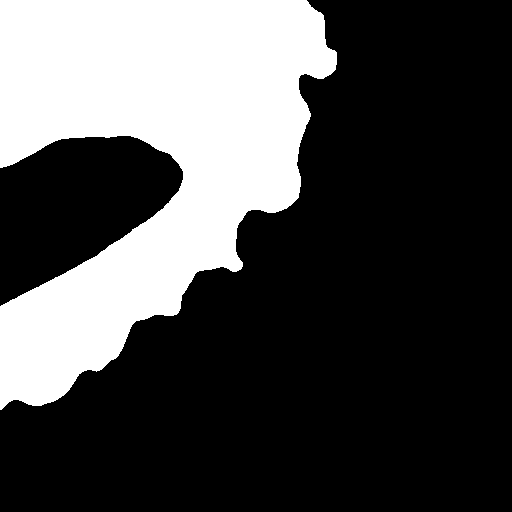}
\end{tabular}
\vspace{-9pt}
\begin{tcolorbox}[colback=gray!20, colframe=black!0, top=0.4pt, bottom=0.4pt]
\scriptsize
Prompt: \textit{Is there a specific natural asset in this locality that plays a dual role in assisting transportation and supporting irrigation for farming?}
\end{tcolorbox}
\vspace{-3pt}
\begin{tabular}{ccc}
\includegraphics[width=0.315\linewidth]{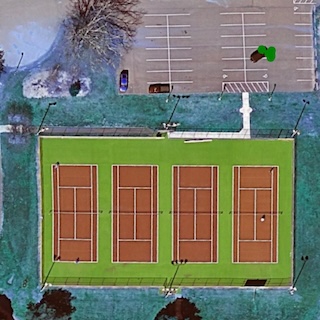} &
\includegraphics[width=0.315\linewidth]{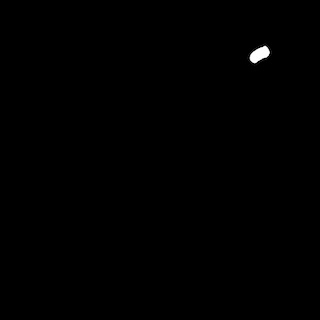} &
\includegraphics[width=0.315\linewidth]{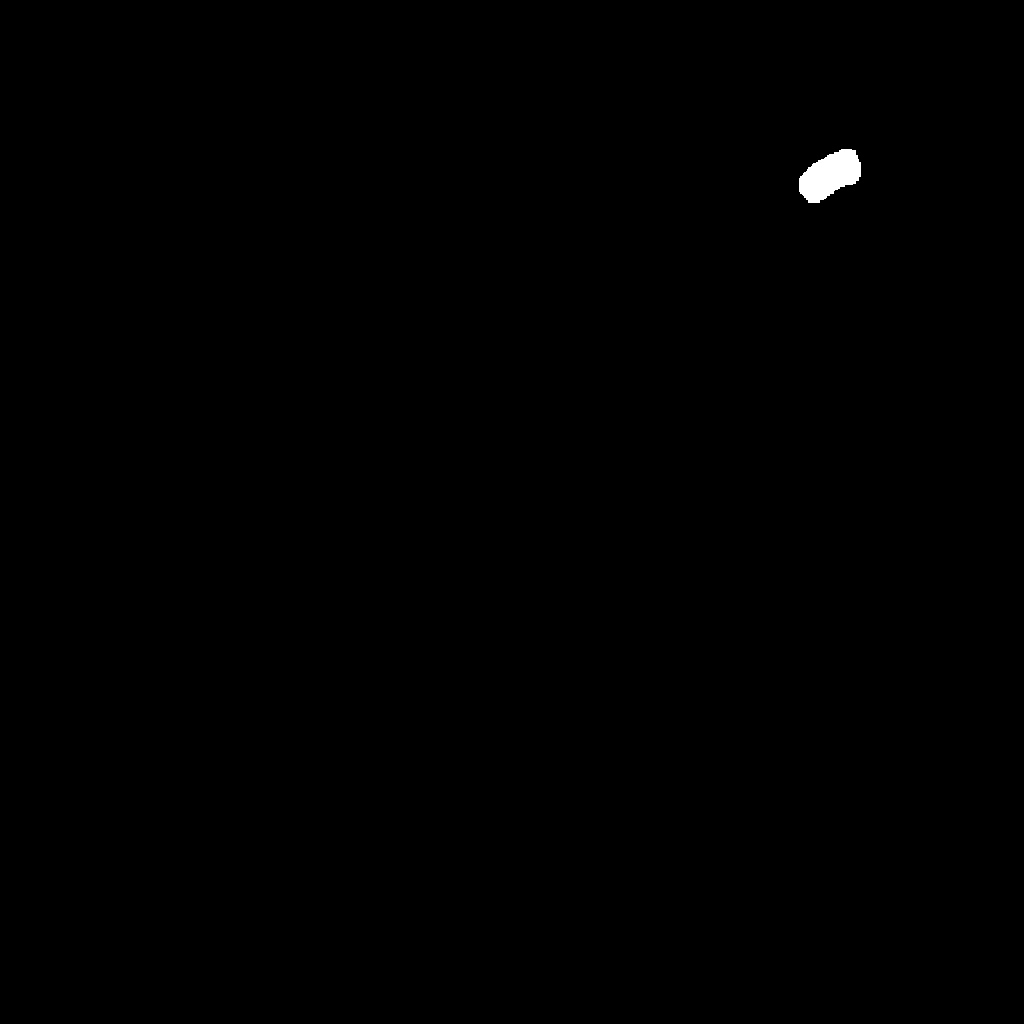}
\end{tabular}
\vspace{-9pt}
\begin{tcolorbox}[colback=gray!20, colframe=black!0, top=0.4pt, bottom=0.4pt]
\scriptsize
Prompt: \textit{The vehicle on the upper right.}
\end{tcolorbox}
\caption{Qualitative results of the LoRA-tuned generative VLM pipeline on reasoning (first and second rows) and referring (third row) tasks using remote sensing datasets.}
\label{fig:reasoning-seg-qual}
\end{figure}

\begin{table*}
\centering
\begin{tabular}{llcccccccc}
\toprule
& \multirow{2}{*}{Method} & \multirow{2}{*}{LLM} & \multicolumn{3}{c}{Trained on RS data} & \multicolumn{2}{c}{RRSIS-D} & \multicolumn{2}{c}{EarthReason} \\
& & & LLM & Mask Decoder & Extra & Val & Test & Val & Test \\
\midrule
\multicolumn{3}{l}{\textit{Zero-shot methods}} \\
& GPT-Image-1 & GPT-5 & \xmark & \xmark & \xmark & 20.1 & 17.2 & 38.4 & 41.0 \\
& \textbf{Ours} & GPT-5 & \xmark & \xmark & \xmark & \textbf{25.8} & \textbf{24.9} & \textbf{46.0} & \textbf{47.4} \\
\midrule
\multicolumn{3}{l}{\textit{Classical methods}} \\
& RRSIS~\cite{yuan2023rrsis} & BERT-base & \cmark & \cmark & \cmark & 60.2 & 59.4 & -- & -- \\
& LAVT~\cite{yang2022lavt} & BERT-base & \cmark & \cmark & \cmark & 61.5 & 61.0 & -- & -- \\
& DiffRIS~\cite{dong2025diffris} & CLIP & \cmark & \cmark & \cmark & 63.6 & 62.2 & -- & -- \\
& FIANet~\cite{lei2024fianet} & BERT-base & \cmark & \cmark & \cmark & -- & 64.0 & -- & -- \\
& RMSIN~\cite{liu2024rotated} & BERT-base & \cmark & \cmark & \cmark & 65.1 & 64.2 & -- & -- \\
& RSRefSeg~\cite{chen2025rsrefseg} & SigLIP-So & \cmark & \cmark & \cmark & -- & 64.7 & -- & -- \\
& SBANet~\cite{li2025sbanet} & BERT-base & \cmark & \cmark & \cmark & 66.7 & 65.5 & -- & -- \\
& BTDNet~\cite{zhang2025btdnet} & BERT-base & \cmark & \cmark & \cmark & 66.9 & 66.0 & -- & -- \\
\midrule
\multicolumn{3}{l}{\textit{Based on generative VLMs}} \\
& NExT-Chat~\cite{zhang2023next} & Vicuna-7B & \cmark & \cmark & \cmark & 27.0 & 25.0 & -- & -- \\
& LISA~\cite{lai2024lisa} & Vicuna-7B & LoRA & \cmark & \cmark & 27.8 & 26.8 & 61.0 & 60.9 \\
& PixelLM~\cite{ren2024pixellm} & Vicuna-7B & LoRA & \cmark & \cmark & 33.9 & 31.7 & 57.9 & 60.0 \\
& PSALM~\cite{zhang2024psalm} & phi-1.5-1.3B & \cmark & \cmark & \cmark & -- & -- & 66.6 & 68.3 \\
& SegEarth-R1~\cite{li2025segearth-r1} & phi-1.5-1.3B & \cmark & \cmark & \cmark & 67.6 & 66.4 & 68.6 & 70.7 \\
& GeoPixel~\cite{shabbir2025geopixel} & InternLM2-7B & LoRA & \cmark & \cmark & 68.0 & 67.3 & -- & -- \\
& \textbf{Ours} & Qwen3-VL-2B & LoRA & \xmark & \xmark & \textbf{68.1} & \textbf{67.6} & \textbf{70.6} & \textbf{72.7} \\
\bottomrule
\end{tabular}
\caption{Results of our generative VLM-based approach for text-based remote sensing segmentation on reasoning and referring tasks. We evaluate on test and validation sets from RRSIS-D~\cite{liu2024rotated} (referring) and EarthReason~\cite{li2025segearth-r1} (reasoning) datasets.}
\label{tab:reasoning-seg-results}
\end{table*}

\vspace{0.5em}
\noindent\textbf{Enabling training-free referring and reasoning segmentation.} We evaluate our generative-VLM based approach for referring and reasoning segmentation tasks on the validation and test splits of RRSIS-D~\cite{liu2024rotated} and EarthReason~\cite{li2025segearth-r1}. The upper part of~\autoref{tab:reasoning-seg-results} shows the results of our training-free approach. In this setup, a proprietary generative VLM is prompted to output click positions, which are fed into SAM (second row). This yields better segmentation results than directly prompting the same VLM to output segmentation masks (first row). However, these promising results still fall short of the current state of the art. This limitation likely arises from the difficulty of current VLMs to perform challenging tasks, such as spatial reasoning and referring segmentation on remote sensing imagery.

\vspace{0.5em}
\noindent\textbf{Achieving SOTA referring and reasoning segmentation with LoRA-tuned generative VLM.} We evaluate the fine-tuned generative VLM-based pipeline on the same datasets and tasks as in the zero-shot setup. As shown in \autoref{tab:reasoning-seg-results}, this fine-tuning proves highly effective, achieving state-of-the-art performance on both referring (RRSIS-D) and reasoning (EarthReason) segmentation tasks. Unlike previous methods that require full training of LLMs, mask decoders, or additional components, our approach avoids heavy retraining. We fine-tune only a lightweight subset of LLM parameters using LoRA, while keeping the mask generator (SAM) frozen. This design resulted efficient, reducing the number of trainable components compared to prior methods without compromising performance.

\vspace{0.5em}
\noindent\textbf{Qualitative results.} \autoref{fig:ovss-main} presents results from our contrastive pipeline for multi-class and single-class OVSS. Due to space limits, we show a subset of datasets: OpenEarthMap~\cite{xia2023openearthmap} and LoveDA~\cite{wang2021loveda} for multi-class, and Inria~\cite{maggiori2017can} for single-class. Our approach correctly identifies most classes, with minor errors in challenging categories (\eg, trees, dense vegetation) and occasional misclassifications in crowded scenes (\eg, buildings vs. roads). For the generative VLM-based approach, \autoref{fig:reasoning-seg-qual} shows examples from EarthReason and RRSIS-D for reasoning and referring segmentation. Each row displays the input image with predicted clicks, predicted masks, and ground truths. As observed in the first-row example, our method localises the correct area even when the main object differs from the question target (\eg, the tennis court in the top-right). Additional examples highlight handling of small objects and complex shapes. However, the approach sometimes struggle with ambiguous descriptions, especially when target involves multiple regions. Moreover, SAM’s limitations can lead to inaccurate masks for non-well-delimited areas. Full visualisations and in-depth analysis are in the Appendix.

\subsection{Ablation Studies}

\noindent\textbf{Effect of SAM scale and grid density on contrastive VLM-based pipeline.} We conduct ablation studies on SAM size and grid clicks using OEM, LoveDA, UAVid (multi-class), and CHN6 (single-class). From the results reported in \autoref{tab:ablations-SAM}, we observe that the largest variant of SAM achieves the highest performance across datasets, which we use as the default. For experiments with different grid sizes, performance improves steadily up to a grid size of $20 \times 20$, after which it plateaus. Consequently, we adopt a $29 \times 29$ grid for the final configuration, as it yields superior metrics across most benchmarks. However, in more computationally constrained setups, it would be possible to adopt smaller SAM without significant degradation on performance.

\begin{table}[h]
\centering
\resizebox{\linewidth}{!}{
\begin{tabular}{lccccc}
\hline
SAM & \# clicks & OEM & LoveDA  & UAVid & CHN6 \\
\hline
SAM-Tiny & 29 $\times$ 29 & 33.9 & 37.7 & 43.8 & 35.2 \\
SAM-Base & 29 $\times$ 29 & 34.2 & 38.1 & 44.2 & 36.2 \\
\midrule
SAM-Large & 10 $\times$ 10 & 29.7 & 36.2 & 40.8 & 31.0 \\
SAM-Large & 20 $\times$ 20 & 33.1 & 38.1 & 44.1 & 35.6 \\
SAM-Large & 29 $\times$ 29 & \textbf{34.2} & \textbf{38.2} & \textbf{44.3} & \textbf{36.4} \\
\bottomrule
\end{tabular}
}
\caption{Ablation of SAM scale and grid density on contrastive VLM-based approach.}
\label{tab:ablations-SAM}
\end{table}

\begin{table}[h]
\centering
\resizebox{\linewidth}{!}{
\begin{tabular}{lccccc}
\toprule
\multirow{2}{*}{Method} & \multirow{2}{*}{\# clicks} & \multicolumn{2}{c}{EarthReason} & \multicolumn{2}{c}{RRSIS-D} \\
&  & Val & Test & Val & Test\\
\midrule
Qwen2.5-VL-7B & 6 & 67.8 & 69.3 & 61.4 & 60.9 \\
Qwen3-VL-4B & 6 & 70.4 & 71.7 & 67.3 & 67.2 \\
\midrule
Qwen3-VL-2B & 2 & 63.8 & 64.9 & 61.1 & 61.0 \\
Qwen3-VL-2B & 4 & 67.9 & 69.8 & 66.4 & 66.4 \\
Qwen3-VL-2B & 6 & \textbf{70.6} & \textbf{72.7} & \textbf{68.1} & \textbf{67.6} \\
\bottomrule
\end{tabular}
}
\caption{Ablation of generative VLM scale and click configuration.}
\label{tab:ablation-reasoning}
\end{table}

\noindent\textbf{Effect of generative VLM scale and click configuration.} We ablate VLM size and number of clicks as depicted in~\autoref{tab:ablation-reasoning}, on EarthReason and RRSIS-D for reasoning and referring segmentation, respectively. According to the results, upgrading from QwenVL-2.5 to QwenVL-3 improves performance, making QwenVL-3 our baseline. Further experiments comparing the 4B ($\sim$80M trainable parameters) and 2B ($\sim$50M trainable parameters) variants of Qwen3-VL show a performance gain for the smaller 2B model, which we then use for the main experiments. In terms of click configuration, performance increases consistently up to six clicks, which improves results by $+6.8/+7.8$ (EarthReason val/test) and $+7.0/+6.6$ (RRSIS-D val/test) over two clicks variant.

\section{Conclusion}
\label{sec-conclusions}
We introduced a simple yet powerful approach for zero-shot text-based segmentation of remote sensing imagery. Our approach combines contrastive (CLIP) and generative (GPT-5, Qwen-VL) VLMs with the Segment Anything Model (SAM). The resulting two pipelines achieve state-of-the-art results on 19 remote sensing benchmarks, including open-vocabulary, referring, and reasoning segmentation. The contrastive pipeline enables fully training-free OVSS, while the generative pipeline supports both zero-shot inference (via GPT-5) and lightweight LoRA fine-tuning (via Qwen-VL) for more complex linguistic reasoning. Despite the used VLMs being primarily pre-trained on natural images, our results demonstrate that the proposed approach remains effective for earth perception tasks. As foundation models continue to evolve, we anticipate even better zero-shot capabilities and improved alignment between visual and textual representations for more complex, real-world geospatial understanding.

\vspace{1em}
\textbf{Acknowledgments.} We thank Rim Sleimi and Nicla Notarangelo for great discussions. This work was supported by FNR HPC BRIDGES project, with reference HPC BRIDGES/2022/17978225/AI4CC. Experiments were performed on MeluXina, special thanks to LuxProvide team for the support.

\appendix
\section*{Appendix}

\section{Implementation Details}
\label{sec:sup-implementation}

\begin{figure*}
\centering \footnotesize \setlength{\tabcolsep}{2pt}
\begin{tabular}{cccccc}
clicks: 1; IoU: 47.7 & clicks: 2; IoU: 25.1 & clicks: 3; IoU: 66.3 & clicks: 4; IoU: 80.1 & clicks: 5; IoU: 88.8 & clicks: 6; IoU: 90.6 \\
\includegraphics[width=0.155\linewidth]{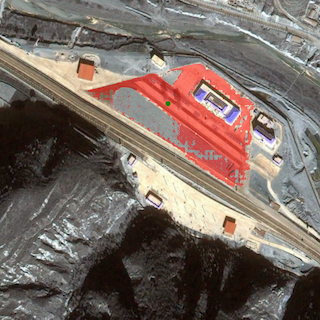} &
\includegraphics[width=0.155\linewidth]{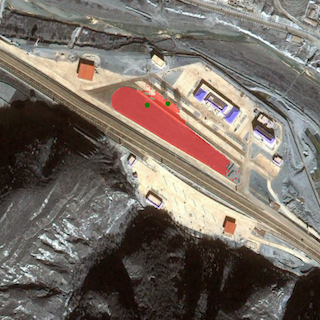} &
\includegraphics[width=0.155\linewidth]{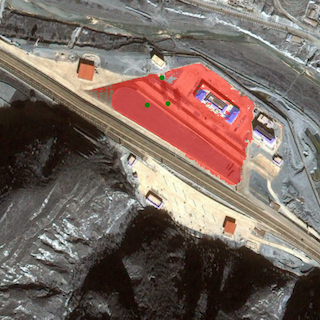} & 
\includegraphics[width=0.155\linewidth]{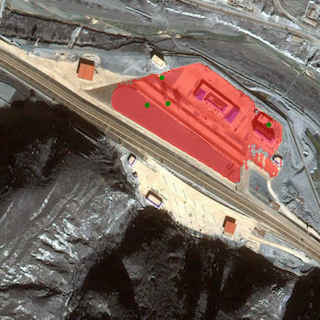} & 
\includegraphics[width=0.155\linewidth]{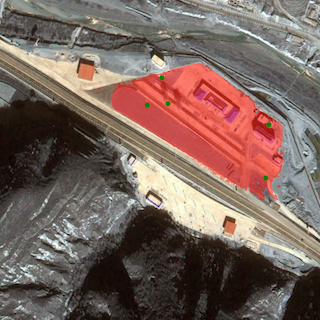} & 
\includegraphics[width=0.155\linewidth]{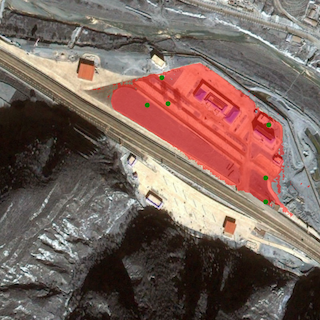} \\
\end{tabular}
\caption{Visualisations from click generation procedure. Masks (red) are produced by SAM prompted by clicks (green). Reported IoU is compared to groung truth mask.}
\label{fig:sup-clicks}
\end{figure*}

\begin{table*}[ht!]
\centering
\resizebox{\textwidth}{!}{
\begin{tabular}{llccccccccc|c}
\toprule
& \multirow{2}{*}{Method} & \multicolumn{4}{c}{Building extraction} & \multicolumn{4}{c}{Road Extraction} & Flood Detection & \multirow{2}{*}{Avg.} \\
& & WHU-A & WHU-S & Inria & xBD-pre & CHN6 & DG & MA & SpaceNet & WBS-SI & \\
\midrule
\multicolumn{3}{l}{\textit{Trained on remote sensing data}} \\
& SegEarth-OV~\cite{li2025segearth} & 49.9 & -- & 48.9 & \textbf{43.1} & 32.8 & \textbf{20.1} & 17.2 & 29.1 & \textbf{57.9} & 37.4 \\
\midrule
\multicolumn{3}{l}{\textit{Zero-shot methods}} \\
& Ours &  \textbf{58.7} & -- & \textbf{53.8} & 39.1 & \textbf{33.5} & 15.0 & \textbf{20.3} & \textbf{29.5} & 56.3 & \textbf{38.3}  \\
\bottomrule
\end{tabular}
}
\caption{Additional results of our contrastive VLM-based approach for text-based remote sensing segmentation on the OVSS task using images of size $896\times896$. Avg. denotes the average across all datasets. Best results are highlighted in \textbf{bold}.}
\label{tab:sup-ovss-896}
\end{table*}

\paragraph{Click generation procedure.}
To fine-tune our generative VLM approach for referring and reasoning segmentation tasks, we use the training splits from the RRSIS-D and EarthReason datasets. Our objective is to train the VLM to output click positions in textual form, which are subsequently used to prompt SAM. For this purpose, we utilise the input images and their corresponding ground truth masks. Each mask $M$ is automatically converted into a sequence of clicks $\mathcal{C}$ without human intervention. 

Inspired by interactive segmentation methods~\cite{sofiiuk2022reviving, antonov2024rclicks}, we adopt an iterative click generation strategy. Formally, at iteration $i$, given the current click sequence $\mathcal{C}_{i-1}$, we compute an intermediate predicted mask using SAM:

\[
    M_{i-1} = \mathscr{S}(I, \text{prompt} = \mathcal{C}_{i-1}).
\]
The discrepancy between $M_{i-1}$ and the ground-truth mask $M$ reveals both under-segmented and over-segmented regions. We define two binary error maps:
\[
    E_{+} = M - M_{i-1}, \quad 
    E_{-} = M_{i-1} - M,
\]
where $E_{+}$ contains pixels that should be included (false negatives), and $E_{-}$ contains pixels that should be excluded (false positives).

Next, we compute a distance transform over the union $E_{+} \cup E_{-}$, which yields a probability distribution emphasizing pixels far from already-correct regions. A new click $c_i$ is then sampled from this distribution:
\[
    c_i \sim \text{DistanceTransform}(E_{+} \cup E_{-}).
\]
If $c_i \in E_{+}$, it is labeled as a \textit{positive} click; if $c_i \in E_{-}$, it is labeled as a \textit{negative} click. The click set is updated as:
\[
    \mathcal{C}_i = \mathcal{C}_{i-1} \cup \{c_i\}.
\]
This process is repeated until a stopping condition is met (\eg, achieving a target IoU or reaching a maximum number of clicks) as depicted in~\autoref{alg:click_generation}. The resulting synthetic click sequences $\mathcal{C}$ are then used to finetune the generative VLM for click generation. \autoref{fig:sup-clicks} illustrates this process.

\begin{algorithm}[t]
\caption{Iterative Click Generation for Synthetic Training Sequences}
\label{alg:click_generation}
\DontPrintSemicolon
\SetAlgoLined

\KwIn{Image $I$, ground-truth mask $M$, SAM model $\mathscr{S}$, maximum iterations $T=6$, IoU threshold $\tau=0.98$}
\KwOut{Click sequence $\mathcal{C}$}

\BlankLine
Initialise click sequence: $\mathcal{C} \leftarrow \emptyset$ \;

\BlankLine
\For{$i \leftarrow 1$ \KwTo $T$}{
    \tcp{Predict intermediate mask from current click set}
    $M_{i-1} \leftarrow \mathscr{S}(I, \text{prompt}=\mathcal{C})$ \;
    
    \tcp{Stop if sufficiently close to ground truth}
    \If{$\text{IoU}(M_{i-1}, M) \ge \tau$}{
        \textbf{break}
    }

    \tcp{Compute false-negative and false-positive regions}
    $E_{+} \leftarrow M - M_{i-1}$ \;
    $E_{-} \leftarrow M_{i-1} - M$ \;

    \tcp{Sample next click using a distance transform over errors}
    $D \leftarrow \text{DistanceTransform}(E_{+} \cup E_{-})$ \;
    Sample $c_i \sim D$ \;

    \tcp{Assign click polarity}
    \uIf{$c_i \in E_{+}$}{
        Label $c_i$ as positive \;
    }
    \Else{
        Label $c_i$ as negative \;
    }

    \tcp{Update click sequence}
    $\mathcal{C} \leftarrow \mathcal{C} \cup \{c_i\}$ \;
}

\BlankLine
\Return{$\mathcal{C}$}
\end{algorithm}

\paragraph{Contrastive VLM inference.}
We use the same CLIP (ViT-B/16) model as in~\cite{li2025segearth-r1}, initialised with the official weights provided by OpenAI. For the text encoder, we adopt the OpenAI ImageNet prompt template, e.g., “a photo of a \textit{{class name}},” as input. Following~\cite{li2025segearth}, we also rename some of the official classes listed on~\autoref{tab:datasets} for OVSS. For CLIP, input images are resized such that the long side is 448 pixels on main paper experiments, and slide inference is performed using a $224 \times 224$ window with a stride of 112. The input to SAM retains the original image dimensions. To avoid memory issues with extremely large images, we cap the maximum image size at 1024 pixels. Images larger than this are split into $1024 \times 1024$ non-overlapping patches, processed individually by SAM, and the resulting mask predictions are then merged.

\begin{table}[h]
\centering
\resizebox{\linewidth}{!}{
\begin{tabular}{llcccc}
\toprule
& Method & VLM & Decoder & Time\textdownarrow & IoU\textuparrow \\
\midrule
\multicolumn{6}{l}{\textit{Contrastive VLM; LoveDA test set}} \\
& SegEarth-OV~\cite{li2025segearth} & 0.15B & SimFeatUp\textsuperscript{$\star$} & 4.1 & 36.9 \\
& Ours & 0.15B & SAM (10$\times$10) & \textbf{3.6} & 36.2 \\
& Ours & 0.15B & SAM (29$\times$29) & 14.0 & \textbf{38.2} \\
\midrule
\multicolumn{6}{l}{\textit{Generative VLM; EarthReason test set}} \\
& SegEarth-R1~\cite{li2025segearth-r1} & 1.3B & Mask2Former\textsuperscript{$\star$} & \textbf{0.44} & 70.7 \\
& Ours & 2B & SAM & 0.58 & \textbf{72.7} \\
\bottomrule
\end{tabular}
}
\caption{Inference time comparison. \textsuperscript{$\star$} indicates that the component is trained on remote sensing data.}
\label{table:inf-time}
\end{table}

\begin{figure*}
\centering \footnotesize \setlength{\tabcolsep}{2pt}
\begin{tabular}{cccc}
Image & Qwen3-VL prompt & GPT-5 & GPT-Image-1 \\
\includegraphics[width=0.08\linewidth]{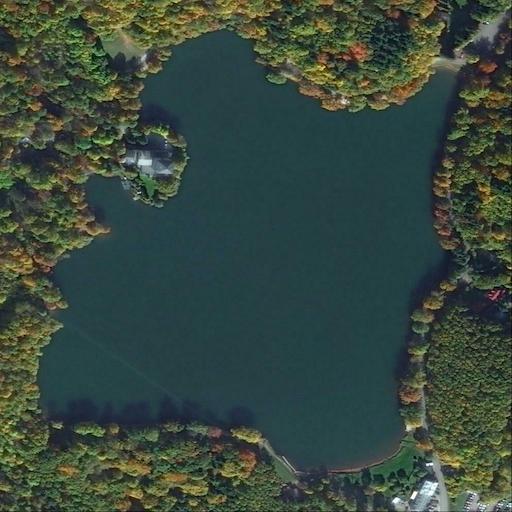} &
\begin{tcolorbox}[width=0.29\linewidth, colback=gray!10, colframe=black!0, left=0.2pt, right=0.2pt, top=0.4pt, bottom=0.4pt]
    \footnotesize
    Given an image and a question, output two sets of coordinates that correspond to click points for segmenting that specific object in the image:
    \begin{enumerate}
        \item \textbf{Positive coordinates:} $(x, y)$ points located inside the object category region.
        \item \textbf{Negative coordinates:} $(x, y)$ points located outside the object region.
    \end{enumerate}
    Rules: Each coordinate must be an integer pair in the format $(x, y)$. The total number of coordinates across both sets must not exceed 6. Each set may contain between 0 and 6 coordinates.  
    Output strictly in this format:
    Positive: [(x1, y1), (x2, y2), ...], Negative: [(x3, y3), (x4, y4), ...]  \\
    \textbf{Question:} Which area depicted in the image would be best suited for a peaceful day of fishing, swimming, or paddle boating?
\end{tcolorbox}
 &
\begin{tcolorbox}[width=0.29\linewidth, colback=gray!10, colframe=black!0, left=0.2pt, right=0.2pt, top=0.4pt, bottom=0.4pt]
    \footnotesize
    Given the input image, find 4 positive clicks covering segmentation mask for the object that answers the \textbf{question}:

    Which area depicted in the image would be best suited for a peaceful day of fishing, swimming, or paddle boating?

    Input image shape is 1024 x 1024.
    Return only click coordinates as json [{'x': ..., 'y': ...}, ...] without any comments. Use reasoning.
\end{tcolorbox} &
\begin{tcolorbox}[width=0.29\linewidth, colback=gray!10, colframe=black!0, left=0.2pt, right=0.2pt, top=0.4pt, bottom=0.4pt]
    Given the input image, create a machine-usable segmentation mask for the object that answers the \textbf{question}:

    Which area depicted in the image would be best suited for a peaceful day of fishing, swimming, or paddle boating?

    The mask must follow these rules:
    - Use a pure black background (RGB 0,0,0).
    - Fill the segmented object with pure white (RGB 255,255,255).
    - Do not include any shading, transparency, or partial blending.
    - Output only the mask image, not the original photo.
\end{tcolorbox} \\
\hline
& \begin{tcolorbox}[width=0.29\linewidth, colback=gray!10, colframe=black!0, left=0.2pt, right=0.2pt, top=0.4pt, bottom=0.4pt]
Positive: [(331, 420), (498, 272)],  \\
Negative: [(12, 55)]
\end{tcolorbox} &
\begin{tcolorbox}[width=0.29\linewidth, colback=gray!10, colframe=black!0, left=0.2pt, right=0.2pt, top=0.4pt, bottom=0.4pt]
    \textless think\textgreater ... \textless /think\textgreater
    
    [\{"x": 120, "y": 760\},
      \{"x": 380, "y": 760\},
      \{"x": 120, "y": 960\},
      \{"x": 380, "y": 960\}]
\end{tcolorbox} &
\includegraphics[width=0.08\linewidth]{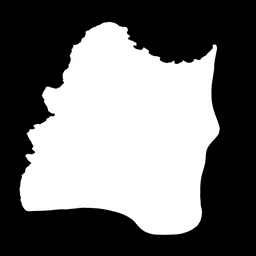}
\end{tabular}
\caption{Top row: input image and corresponding prompts for the different generative VLM settings. Bottom row: outputs produced by each model.}
\label{fig:sup-prompts}
\end{figure*}

\paragraph{Generative VLM inference.}
Examples of prompts used for the generative VLMs (Qwen3-VL, GPT-5, and GPT-Image-1) are provided in ~\autoref{fig:sup-prompts}. The prompts remain fixed across all experiments, with only the input image and the question component varying. Note that Qwen3-VL and GPT-5 produce outputs in textual form, whereas GPT-Image-1 directly generates the corresponding segmentation mask.

\paragraph{Inference time.}
As shown in \autoref{table:inf-time}, our contrastive VLM pipeline is faster than SegEarth-OV~\cite{li2025segearth} while achieving comparable accuracy with a small SAM grid (10$\times$10). Note that larger grids improve IoU at the cost of speed. The generative VLM pipeline is slightly slower than SegEarth-R1~\cite{li2025segearth-r1}, mainly due to the larger 2B-parameter Qwen3-VL model compared to the 1.3B-parameter Phi model, but yields a $+2\%$ IoU improvement. Notably, both SegEarth-OV and SegEarth-R1 rely on mask decoders trained on remote sensing data, whereas our approach does not.

\begin{figure}
\centering \footnotesize \setlength{\tabcolsep}{2pt}
\begin{tabular}{cccc}
Input image & SegEarth-OV~\cite{li2025segearth} & Ours & Ground truth \\
\includegraphics[width=0.232\linewidth]{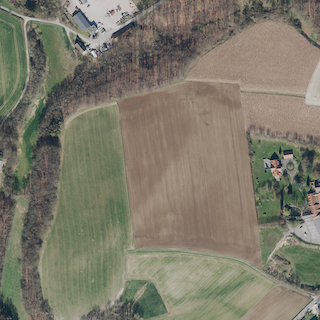} &
\includegraphics[width=0.232\linewidth]{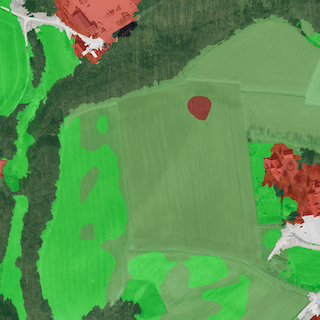} &
\includegraphics[width=0.232\linewidth]{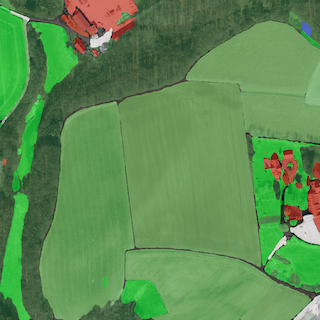} &
\includegraphics[width=0.232\linewidth]{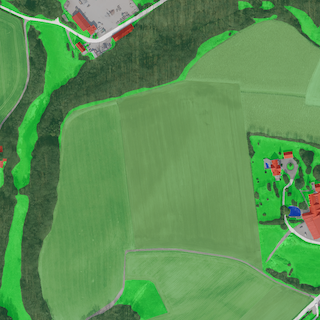} \\
\includegraphics[width=0.232\linewidth]{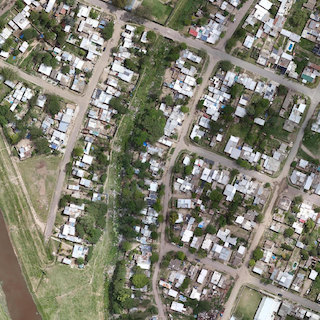} &
\includegraphics[width=0.232\linewidth]{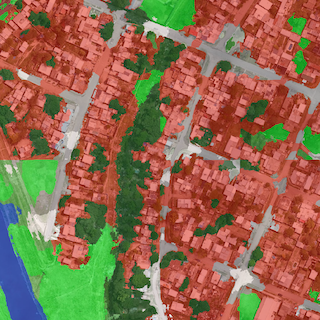} &
\includegraphics[width=0.232\linewidth]{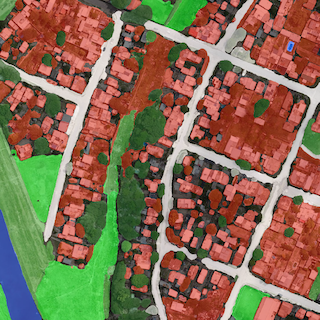} &
\includegraphics[width=0.232\linewidth]{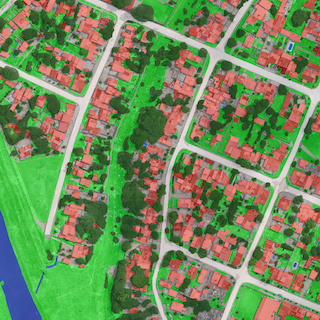} \\
\includegraphics[width=0.232\linewidth]{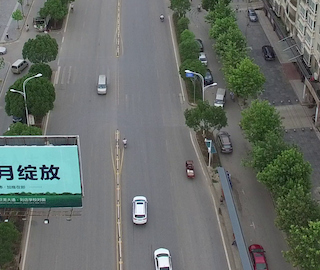} &
\includegraphics[width=0.232\linewidth]{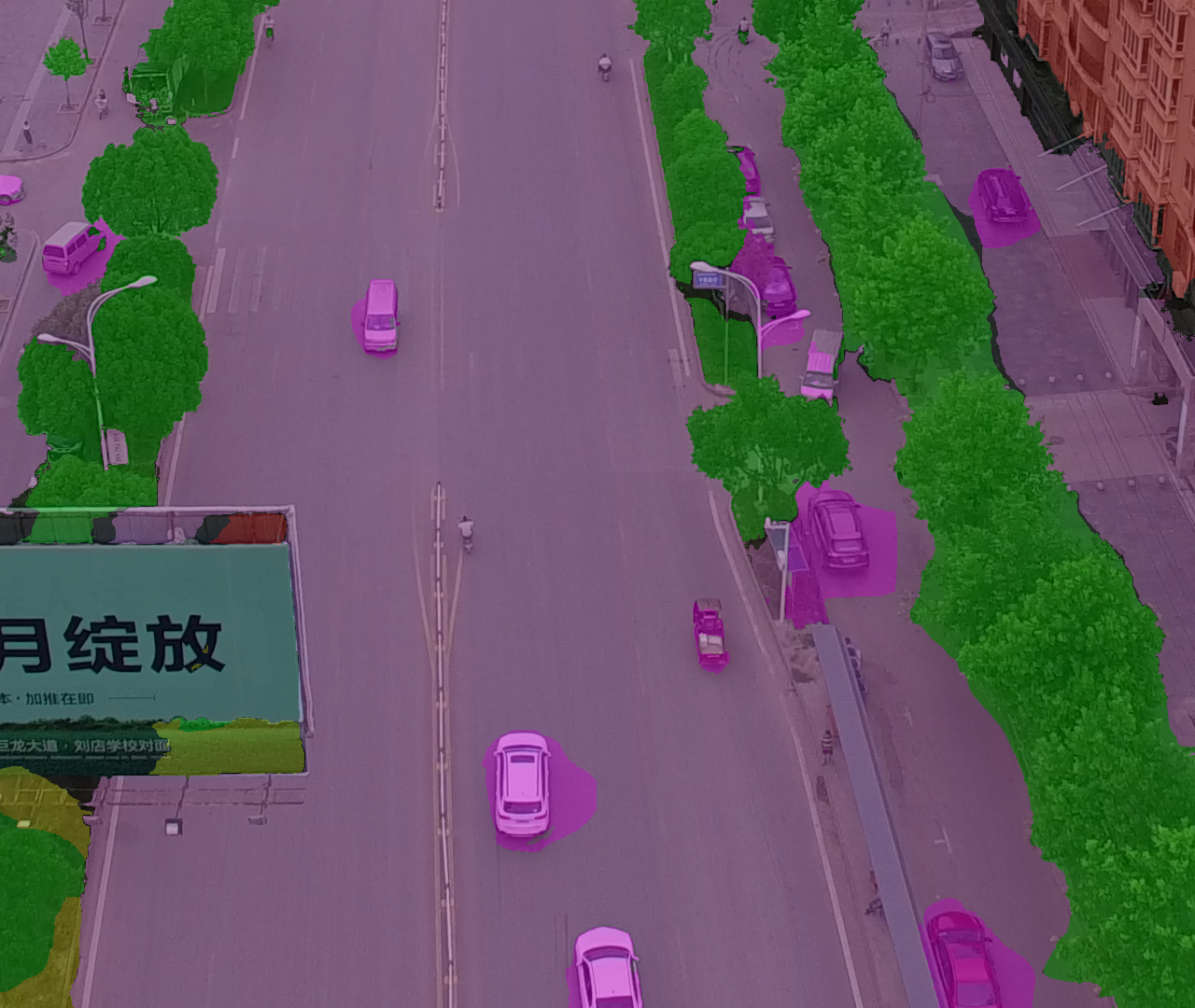} &
\includegraphics[width=0.232\linewidth]{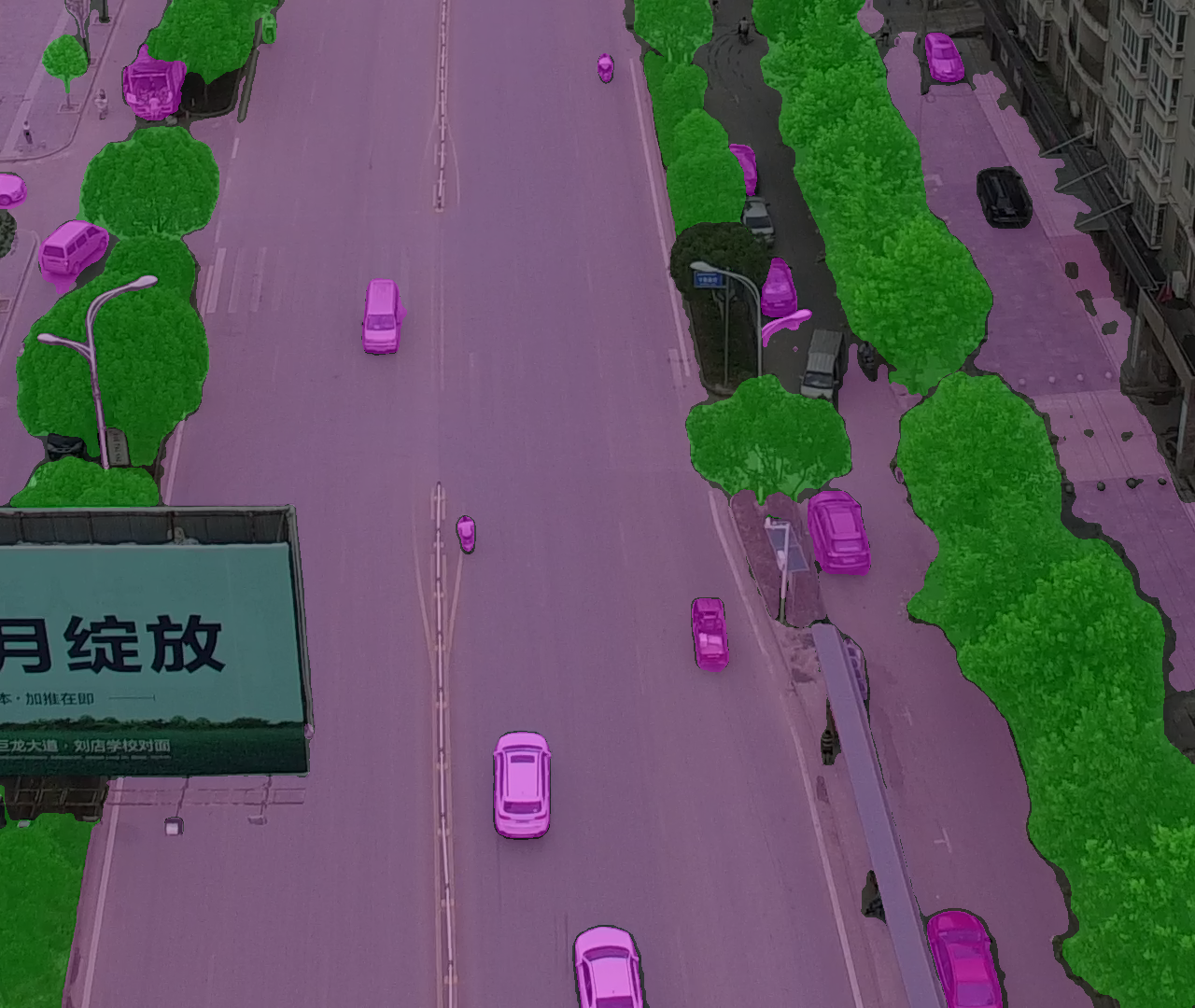} &
\includegraphics[width=0.232\linewidth]{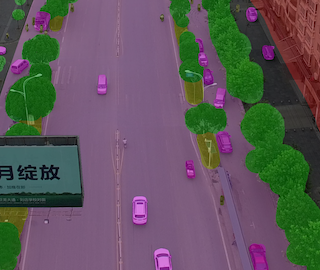} \\
\includegraphics[width=0.232\linewidth]{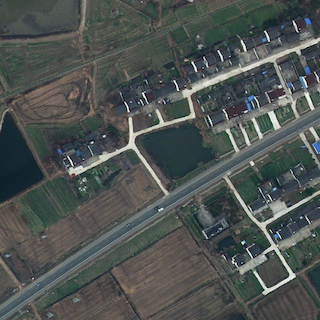} &
\includegraphics[width=0.232\linewidth]{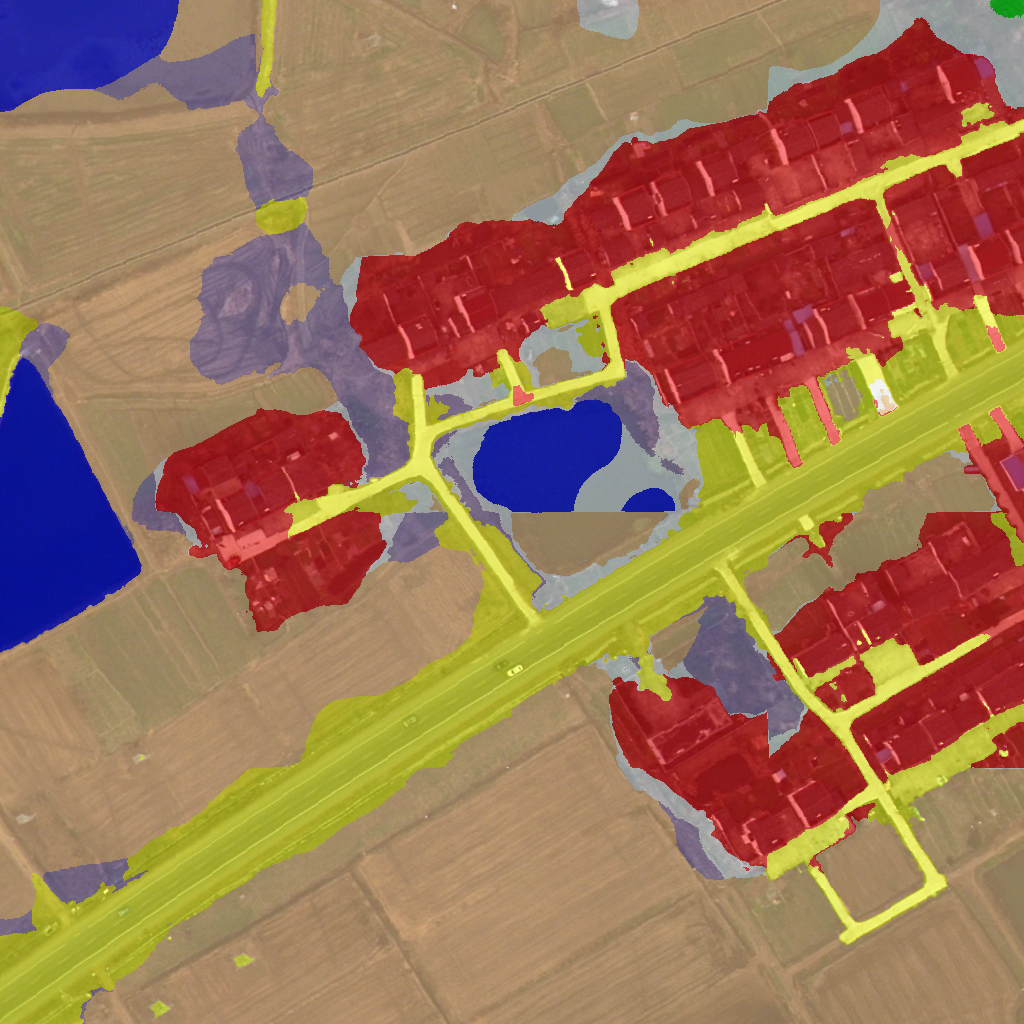} &
\includegraphics[width=0.232\linewidth]{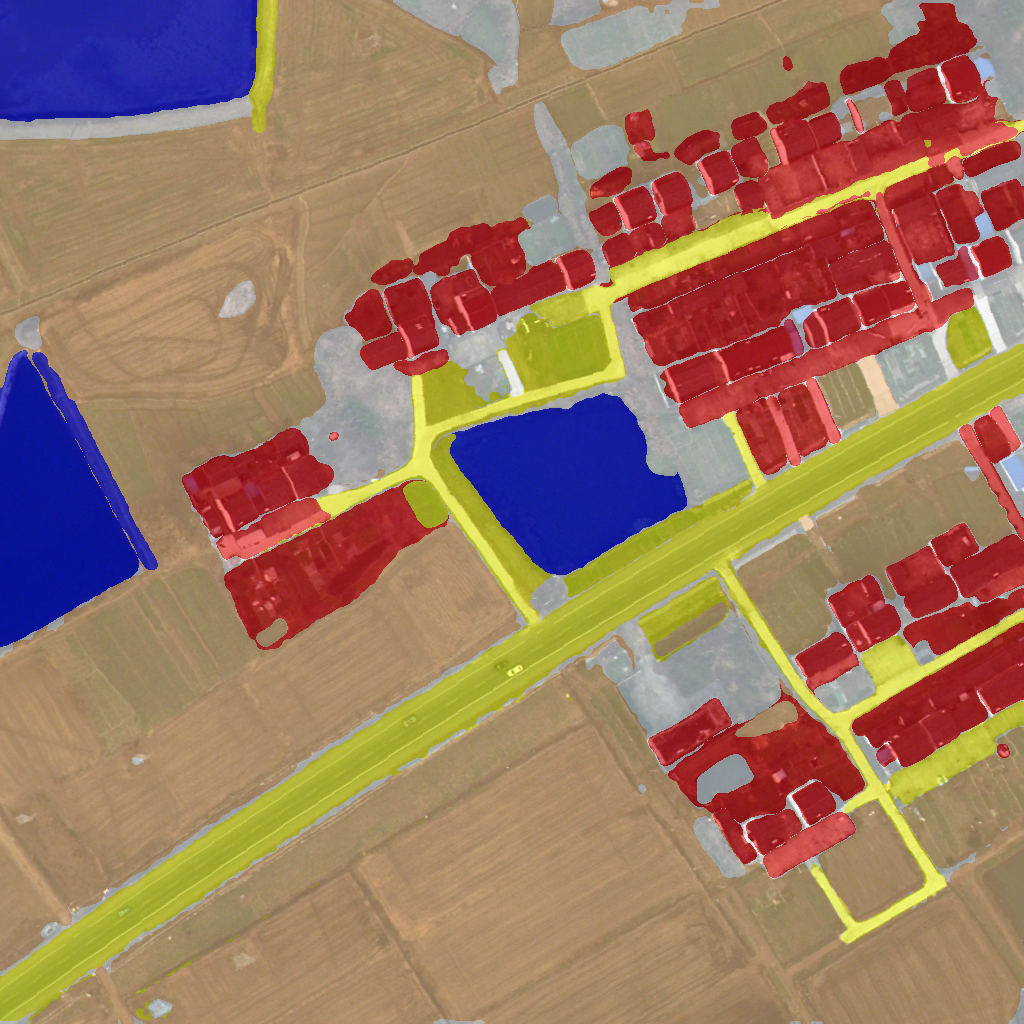} &
\includegraphics[width=0.232\linewidth]{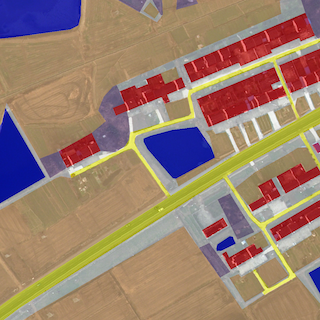} \\
\includegraphics[width=0.232\linewidth]{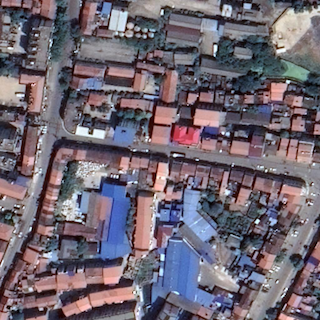} &
\includegraphics[width=0.232\linewidth]{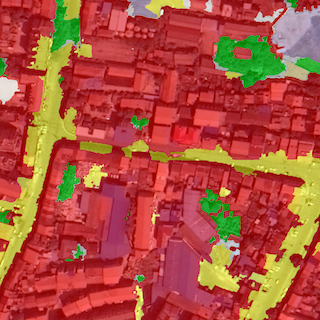} &
\includegraphics[width=0.232\linewidth]{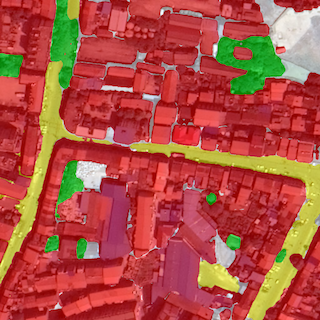} &
\includegraphics[width=0.232\linewidth]{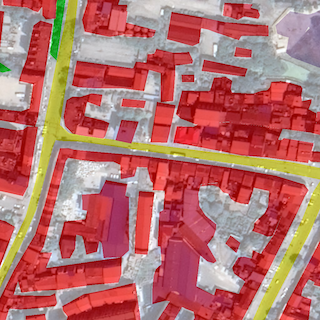}
\end{tabular}
\caption{Qualitative comparison with baseline SegEarth-OV on OVSS datasets.}
\label{fig:sup-ovss-comparison}
\end{figure}

\section{Quantitative Results}
\label{sec:sup-quantitative}

~\autoref{tab:sup-ovss-896} provides additional OVSS results for the single-class extraction datasets using CLIP with an input resolution of $896\times896$. All other experimental settings follow the configuration described in the main experiments. Notably, our model achieves higher average performance than the SegEarth-OV~\cite{li2025segearth} baseline on this setting, despite not being trained on remote sensing data.

\section{Qualitative Results}
\label{sec:sup-qualitative}

\autoref{fig:sup-ovss-comparison} presents additional visualisations of our contrastive VLM-based approach on OVSS. We showcase examples from a subset of multi-class datasets~\cite{xia2023openearthmap,wang2021loveda,lyu2020uavid} and visually compare our predictions with SegEarth-OV~\cite{li2025segearth} and the corresponding ground-truth annotations. Overall, our method produces more precise and better-defined segmentation masks in several categories compared to~\cite{li2025segearth}. This improvement is particularly noticeable for roads (rows 2, 4, and 5) and cars (row 3). In row 2, our approach handles crowded scenes effectively, correctly segmenting roads, buildings, and vegetation. However, small regions between densely packed buildings remain challenging and still cause occasional confusion.

\begin{figure*}
\centering \footnotesize \setlength{\tabcolsep}{2pt}
\begin{tabular}{ccccc}
Input image & SegEarth-R1~\cite{li2025segearth-r1} & GPT-Image-1~\cite{openai_image_generation_2025} & Ours & Ground truth \\
\includegraphics[width=0.19\linewidth]{figures/sup-more-vis/2638.jpg} &
\includegraphics[width=0.19\linewidth]{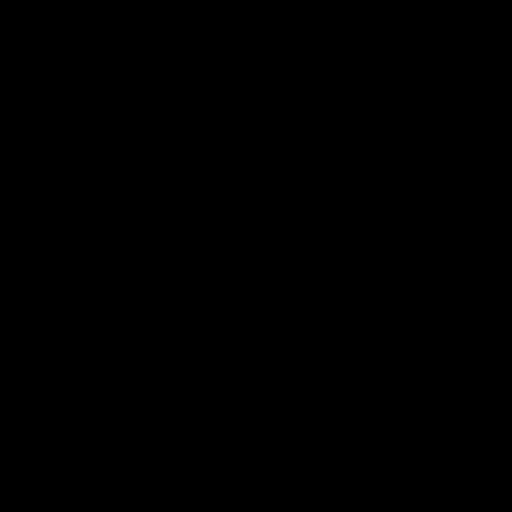} &
\includegraphics[width=0.19\linewidth]{figures/sup-more-vis/2638_gpt-image-1.png} &
\includegraphics[width=0.19\linewidth]{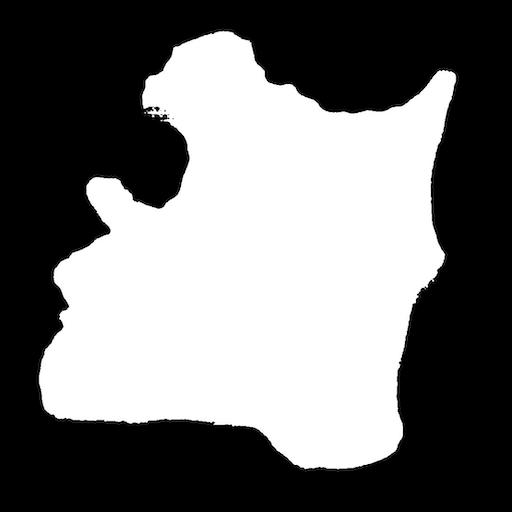} &
\includegraphics[width=0.19\linewidth]{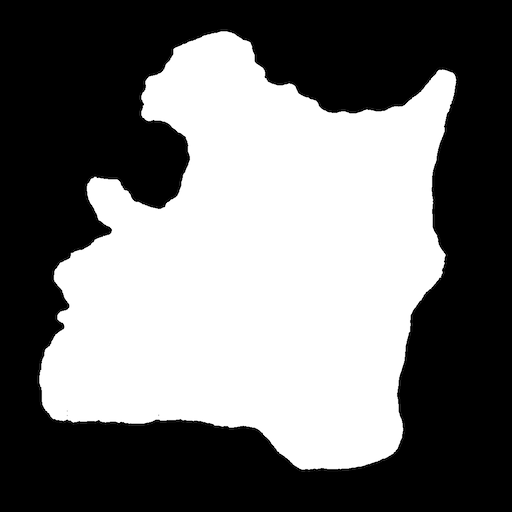}
\end{tabular}
\vspace{-5pt}
\begin{tcolorbox}[colback=gray!20, colframe=black!0, top=0.4pt, bottom=0.4pt, width=0.986\linewidth]
\scriptsize Prompt: \textit{Which area depicted in the image would be best suited for a peaceful day of fishing, swimming, or paddle boating?}
\end{tcolorbox}
\vspace{-3pt}
\begin{tabular}{ccccc}
\includegraphics[width=0.19\linewidth]{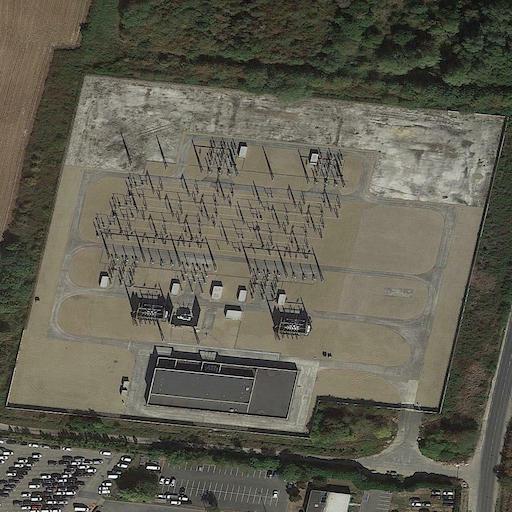} &
\includegraphics[width=0.19\linewidth]{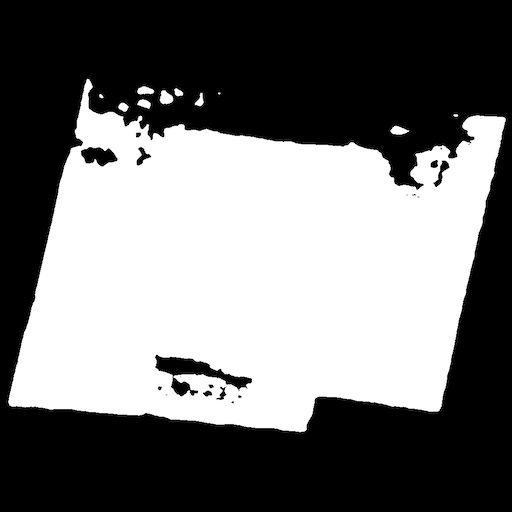} &
\includegraphics[width=0.19\linewidth]{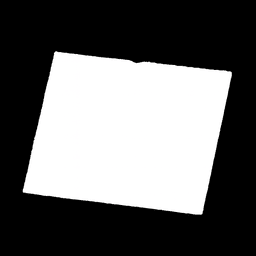} &
\includegraphics[width=0.19\linewidth]{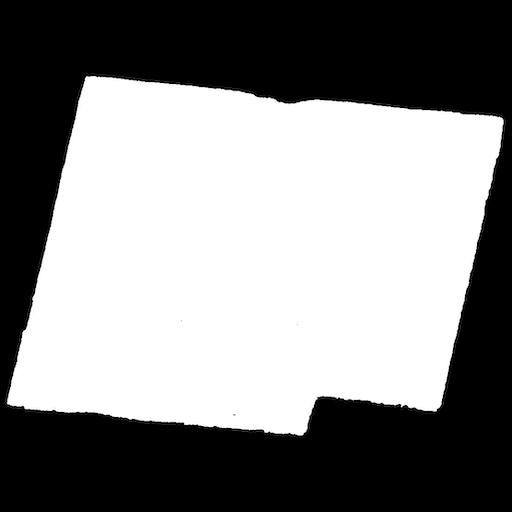} &
\includegraphics[width=0.19\linewidth]{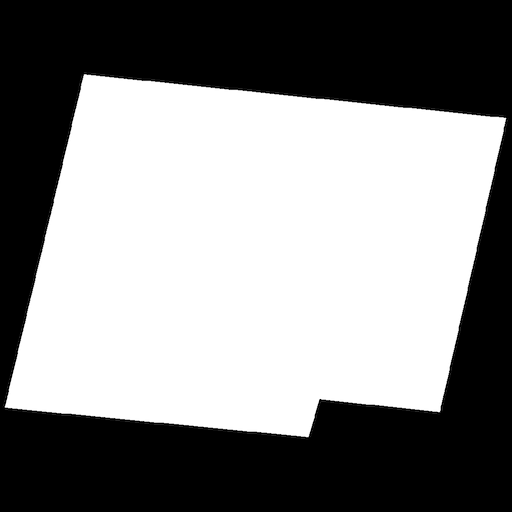}
\end{tabular}
\vspace{-5pt}
\begin{tcolorbox}[colback=gray!20, colframe=black!0, top=0.4pt, bottom=0.4pt, width=0.99\linewidth]
\scriptsize Prompt: \textit{To achieve the best possible electricity delivery in a city, which infrastructure system would you consider the most effective?}
\end{tcolorbox}
\vspace{-3pt}
\begin{tabular}{ccccc}
\includegraphics[width=0.19\linewidth]{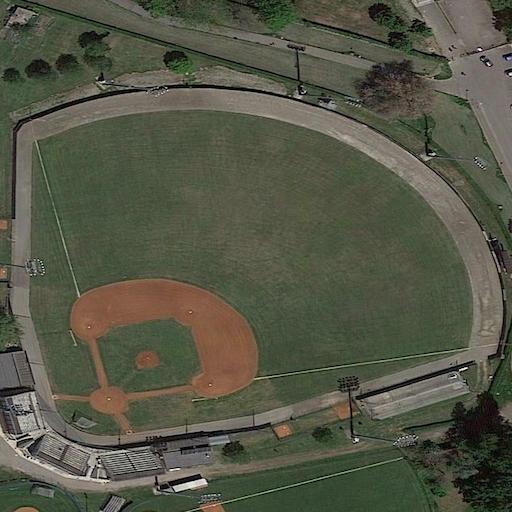} &
\includegraphics[width=0.19\linewidth]{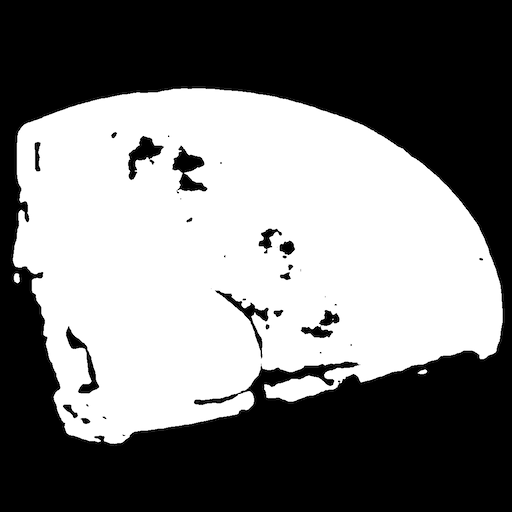} &
\includegraphics[width=0.19\linewidth]{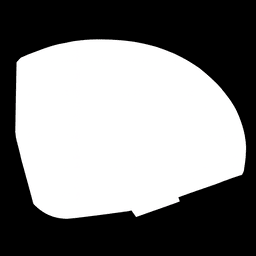} &
\includegraphics[width=0.19\linewidth]{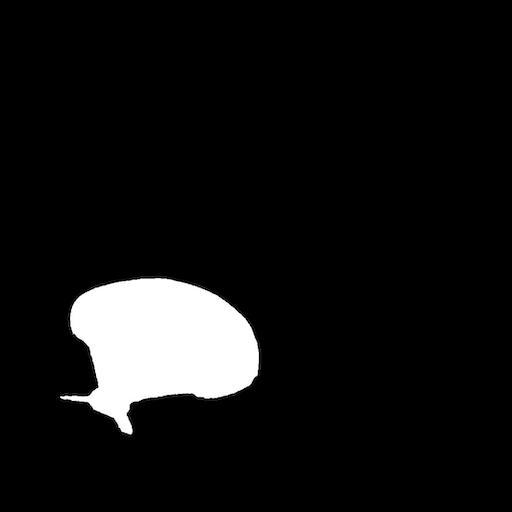} &
\includegraphics[width=0.19\linewidth]{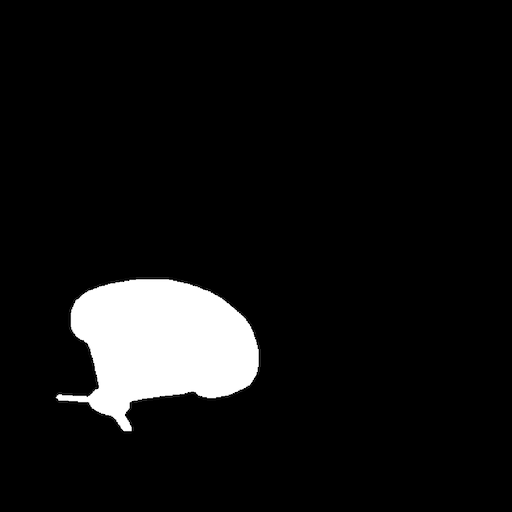}
\end{tabular}
\vspace{-5pt}
\begin{tcolorbox}[colback=gray!20, colframe=black!0, top=0.4pt, bottom=0.4pt, width=0.99\linewidth]
\scriptsize Prompt: \textit{What would be an ideal venue offering ample open areas and adequate seating arrangements for a community gathering or event?}
\end{tcolorbox}
\vspace{-3pt}
\begin{tabular}{ccccc}
\includegraphics[width=0.19\linewidth]{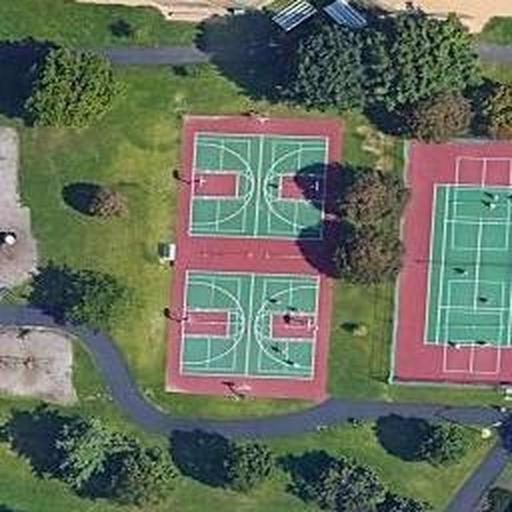} &
\includegraphics[width=0.19\linewidth]{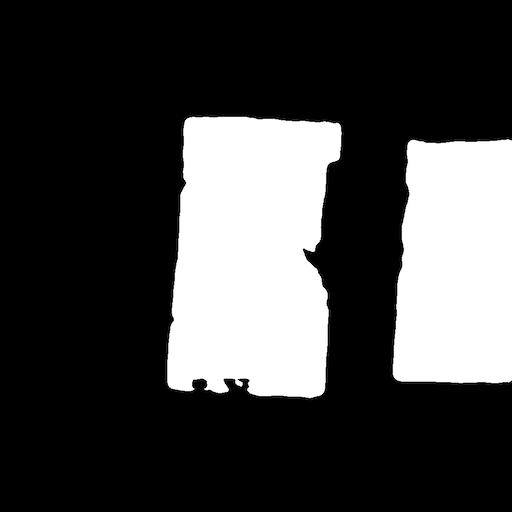} &
\includegraphics[width=0.19\linewidth]{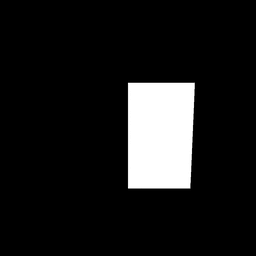} &
\includegraphics[width=0.19\linewidth]{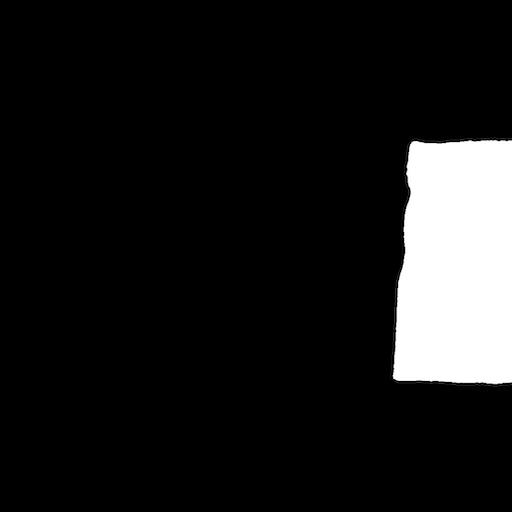} &
\includegraphics[width=0.19\linewidth]{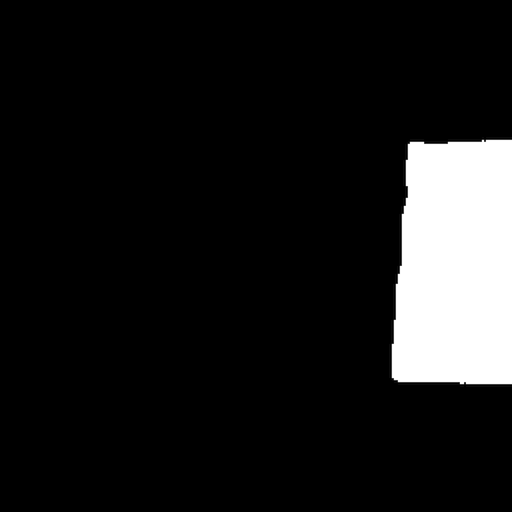}
\end{tabular}
\vspace{-5pt}
\begin{tcolorbox}[colback=gray!20, colframe=black!0, top=0.4pt, bottom=0.4pt, width=0.99\linewidth]
\scriptsize Prompt: \textit{If a person is eager to play racket sports, which specifically designed recreational facility would ensure the most enjoyable experience?}
\end{tcolorbox}
\vspace{-3pt}
\caption{Qualitative comparison with baseline methods on EarthReason dataset.}
\label{fig:sup-reasoning-comparison}
\end{figure*}

\begin{figure}
\centering \footnotesize \setlength{\tabcolsep}{2pt}
\begin{tabular}{cccc}
Input image & Predicted clicks & Predicted mask & Ground truth \\
\includegraphics[width=0.232\linewidth]{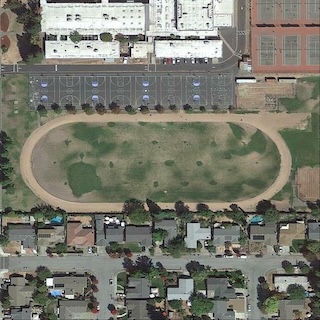} &
\includegraphics[width=0.232\linewidth]{figures/qual-reasoning/98.82_4659_clicks.jpg} &
\includegraphics[width=0.232\linewidth]{figures/qual-reasoning/98.82_4659_predicted_mask.jpg} &
\includegraphics[width=0.232\linewidth]{figures/qual-reasoning/4659.png}
\end{tabular}
\vspace{-9pt}
\begin{tcolorbox}[colback=gray!20, colframe=black!0, top=0.4pt, bottom=0.4pt, width=0.986\linewidth]
\scriptsize Prompt: \textit{Should you wish to improve your overhead serve and join a doubles match, what specific location in the sports complex would you select?}
\end{tcolorbox}
\vspace{-3pt}
\begin{tabular}{cccc}
\includegraphics[width=0.232\linewidth]{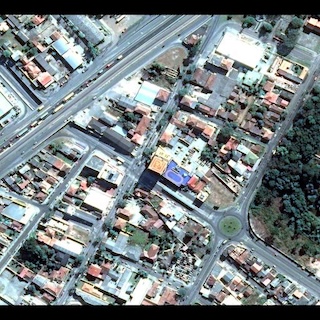} &
\includegraphics[width=0.232\linewidth]{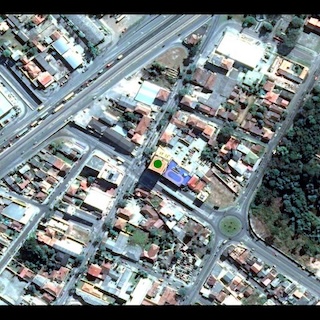} &
\includegraphics[width=0.232\linewidth]{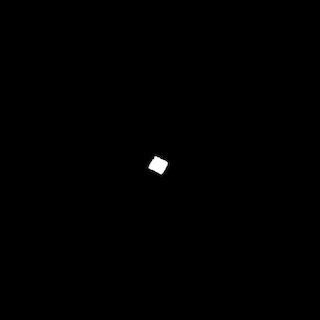} &
\includegraphics[width=0.232\linewidth]{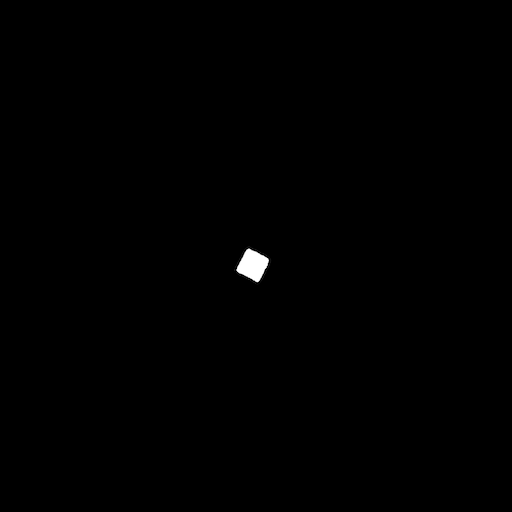}
\end{tabular}
\vspace{-9pt}
\begin{tcolorbox}[colback=gray!20, colframe=black!0, top=0.4pt, bottom=0.4pt, width=0.986\linewidth]
\scriptsize Prompt: \textit{What type of location is commonly designated for the vertical launch and landing of helicopters used in medical emergencies?}
\end{tcolorbox}
\vspace{-3pt}
\begin{tabular}{cccc}
\includegraphics[width=0.232\linewidth]{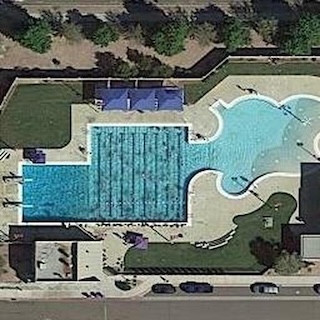} &
\includegraphics[width=0.232\linewidth]{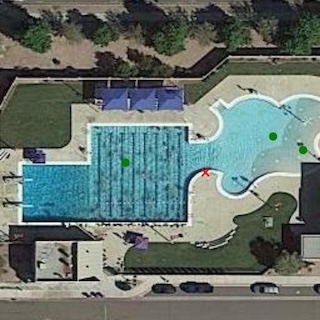} &
\includegraphics[width=0.232\linewidth]{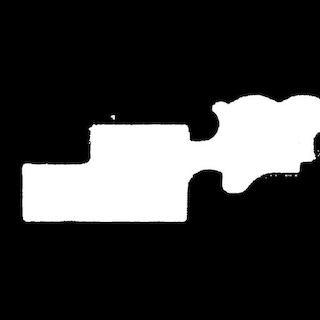} &
\includegraphics[width=0.232\linewidth]{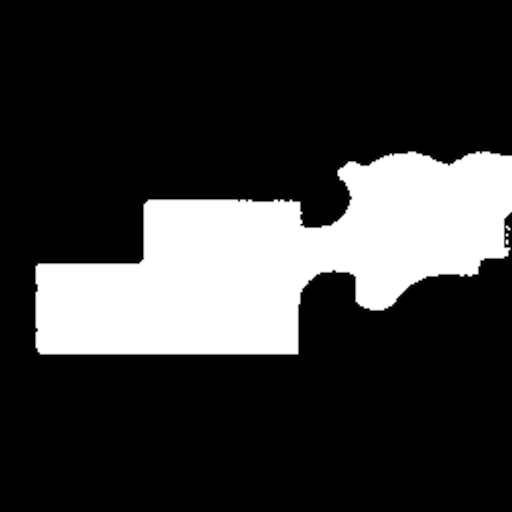}
\end{tabular}
\vspace{-9pt}
\begin{tcolorbox}[colback=gray!20, colframe=black!0, top=0.4pt, bottom=0.4pt, width=0.986\linewidth]
\scriptsize Prompt: \textit{Where might someone find venues for engaging in water-based recreation and sports, which also have facilities for relaxation and soaking up the sun?}
\end{tcolorbox}
\vspace{-3pt}
\begin{tabular}{cccc}
\includegraphics[width=0.232\linewidth]{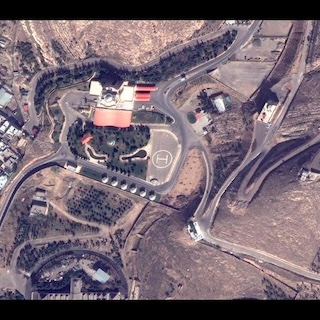} &
\includegraphics[width=0.232\linewidth]{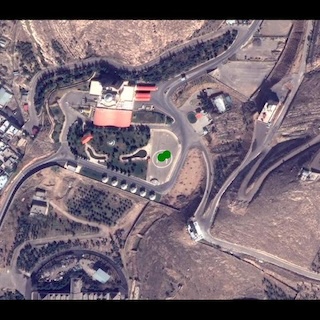} &
\includegraphics[width=0.232\linewidth]{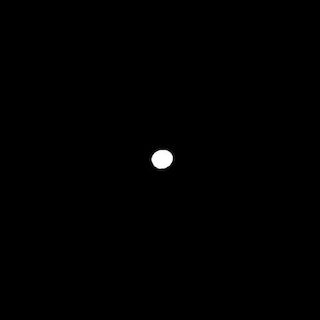} &
\includegraphics[width=0.232\linewidth]{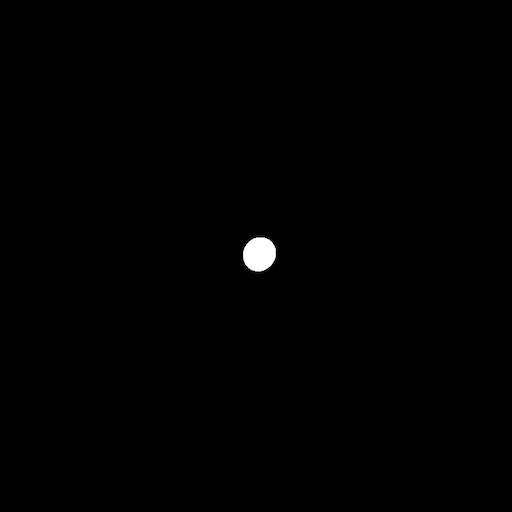}
\end{tabular}
\vspace{-9pt}
\begin{tcolorbox}[colback=gray!20, colframe=black!0, top=0.4pt, bottom=0.4pt, width=0.986\linewidth]
\scriptsize Prompt: \textit{Which facilities in this vicinity specifically cater to the needs of air medical evacuations during emergencies?}
\end{tcolorbox}
\vspace{-3pt}
\begin{tabular}{cccc}
\includegraphics[width=0.232\linewidth]{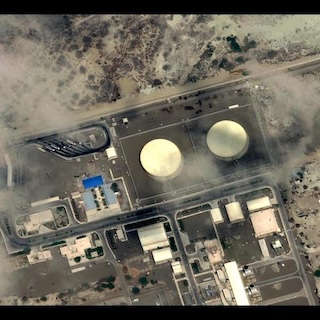} &
\includegraphics[width=0.232\linewidth]{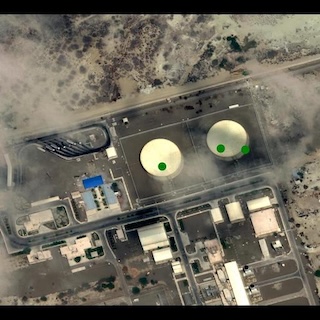} &
\includegraphics[width=0.232\linewidth]{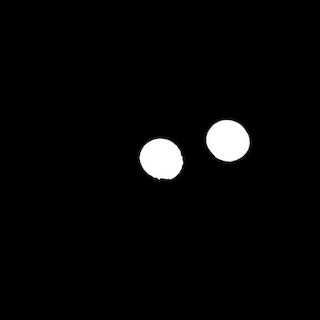} &
\includegraphics[width=0.232\linewidth]{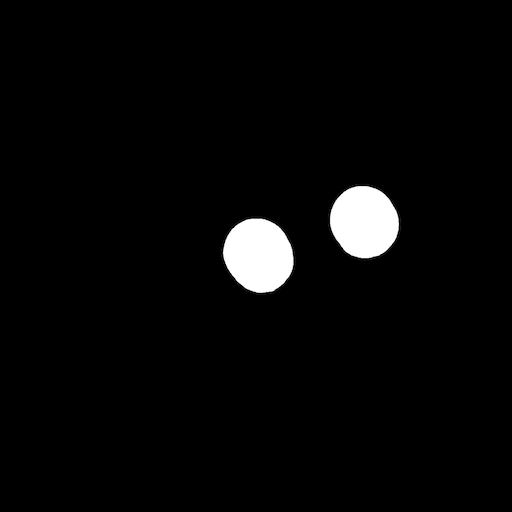}
\end{tabular}
\vspace{-9pt}
\begin{tcolorbox}[colback=gray!20, colframe=black!0, top=0.4pt, bottom=0.4pt, width=0.986\linewidth]
\scriptsize Prompt: \textit{In the scenario of a fire risk, which visible structures in the image require precautionary actions to stop dangerous substance combustions?}
\end{tcolorbox}
\vspace{-3pt}
\caption{Qualitative results for reasoning segmentation on EarthReason dataset.}
\label{fig:sup-reasoning-qualitative}
\end{figure}

\begin{figure}
\centering \footnotesize \setlength{\tabcolsep}{2pt}
\begin{tabular}{cccc}
Input image & Predicted clicks & Predicted mask & Ground truth \\
\includegraphics[width=0.232\linewidth]{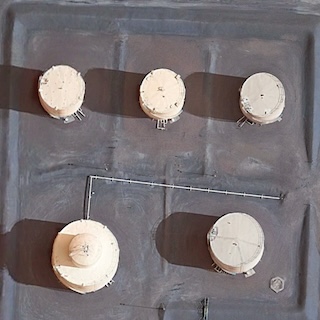} &
\includegraphics[width=0.232\linewidth]{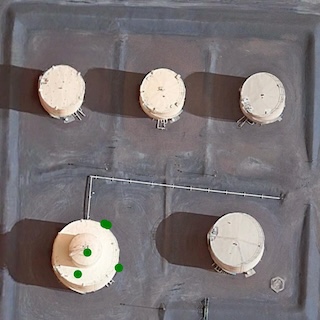} &
\includegraphics[width=0.232\linewidth]{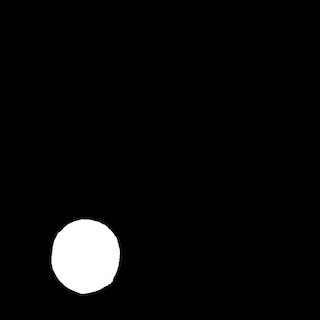} &
\includegraphics[width=0.232\linewidth]{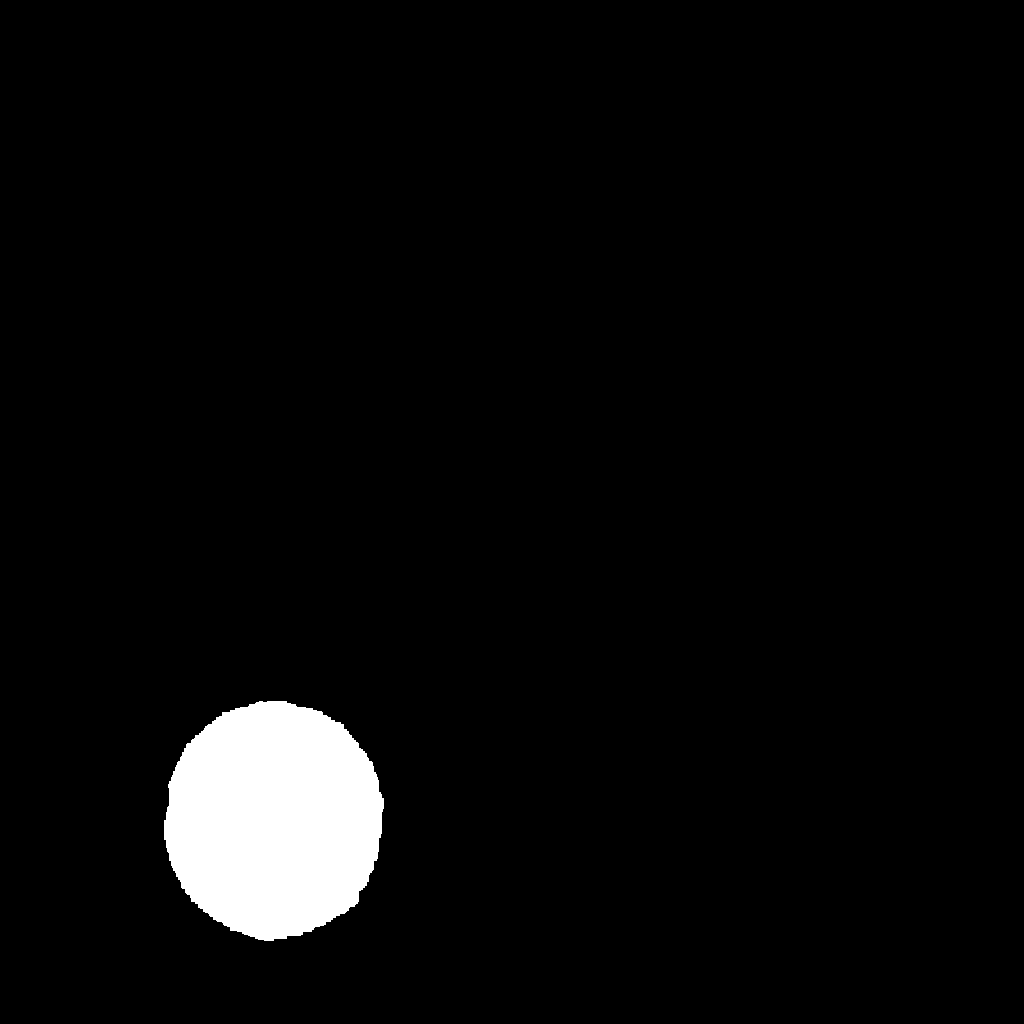}
\end{tabular}
\vspace{-9pt}
\begin{tcolorbox}[colback=gray!20, colframe=black!0, top=0.4pt, bottom=0.4pt, width=0.986\linewidth]
\scriptsize Prompt: \textit{The storage tank on the lower left.}
\end{tcolorbox}
\vspace{-3pt}
\begin{tabular}{cccc}
\includegraphics[width=0.232\linewidth]{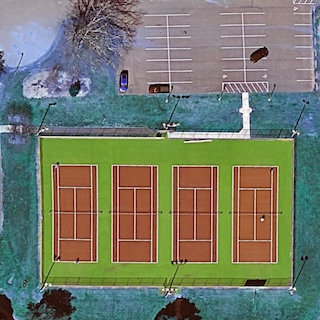} &
\includegraphics[width=0.232\linewidth]{figures/reasoning-main/94.53_6648_clicks.jpg} &
\includegraphics[width=0.232\linewidth]{figures/reasoning-main/94.53_6648_predicted_mask.jpg} &
\includegraphics[width=0.232\linewidth]{figures/reasoning-main/6648.png}
\end{tabular}
\vspace{-9pt}
\begin{tcolorbox}[colback=gray!20, colframe=black!0, top=0.4pt, bottom=0.4pt, width=0.986\linewidth]
\scriptsize Prompt: \textit{The vehicle on the upper right.}
\end{tcolorbox}
\vspace{-3pt}
\begin{tabular}{cccc}
\includegraphics[width=0.232\linewidth]{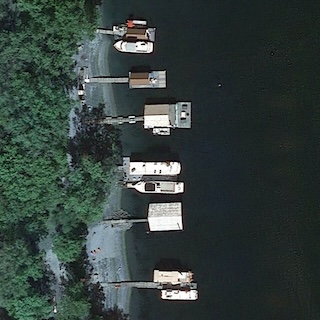} &
\includegraphics[width=0.232\linewidth]{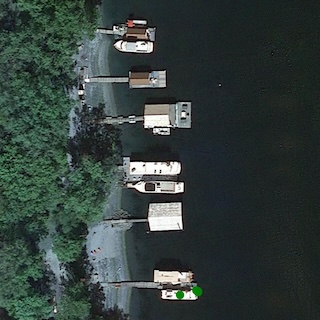} &
\includegraphics[width=0.232\linewidth]{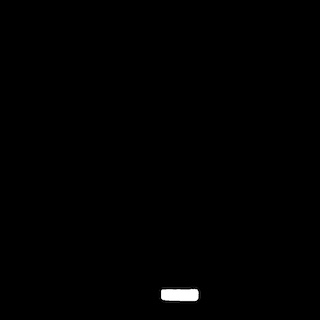} &
\includegraphics[width=0.232\linewidth]{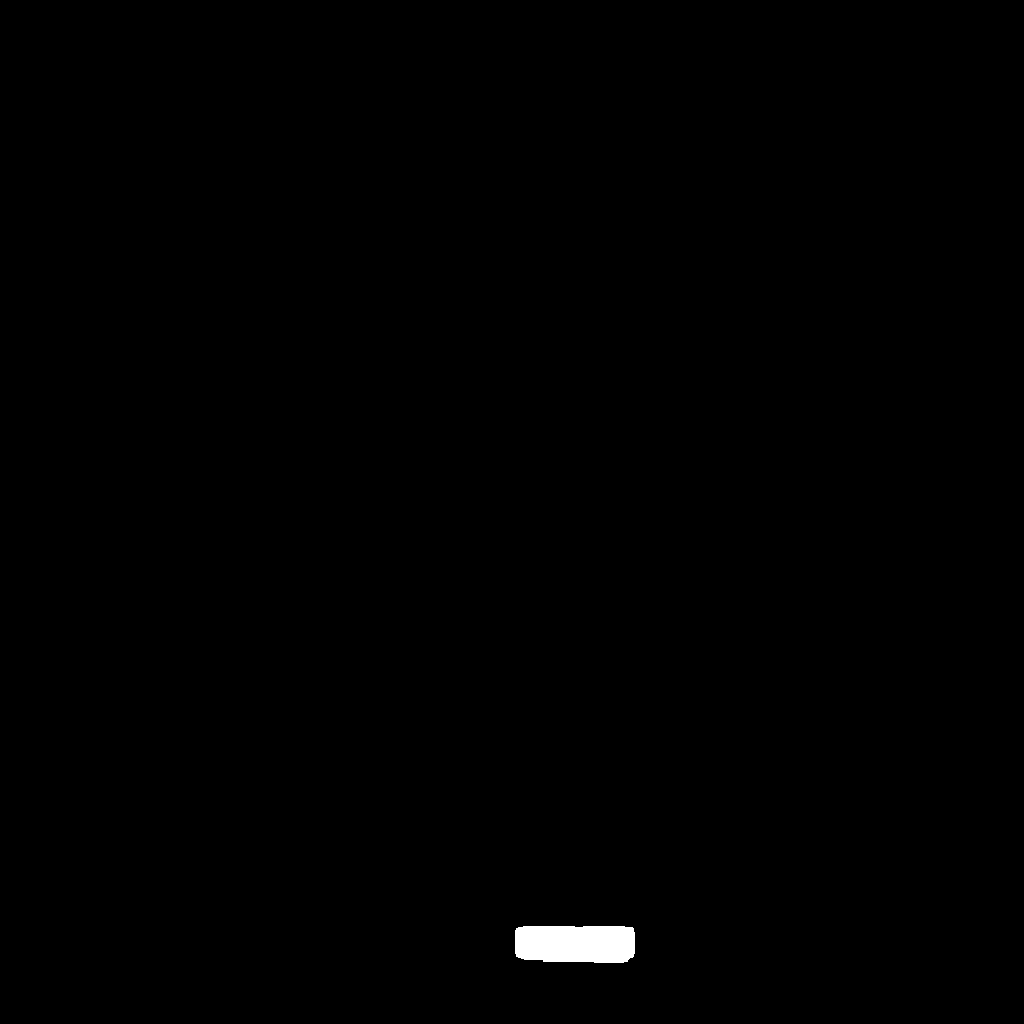}
\end{tabular}
\vspace{-9pt}
\begin{tcolorbox}[colback=gray!20, colframe=black!0, top=0.4pt, bottom=0.4pt, width=0.986\linewidth]
\scriptsize Prompt: \textit{A ship at the bottom.}
\end{tcolorbox}
\vspace{-3pt}
\begin{tabular}{cccc}
\includegraphics[width=0.232\linewidth]{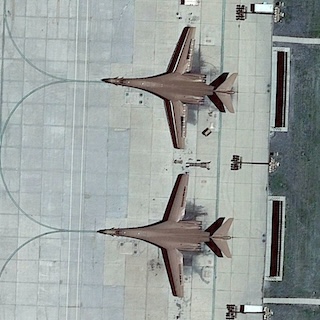} &
\includegraphics[width=0.232\linewidth]{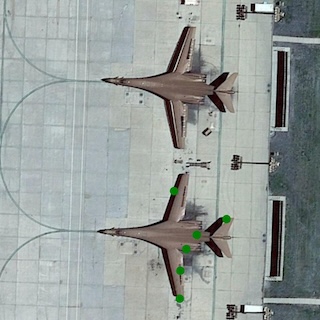} &
\includegraphics[width=0.232\linewidth]{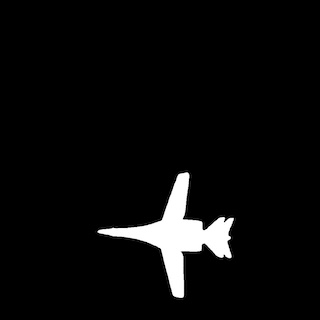} &
\includegraphics[width=0.232\linewidth]{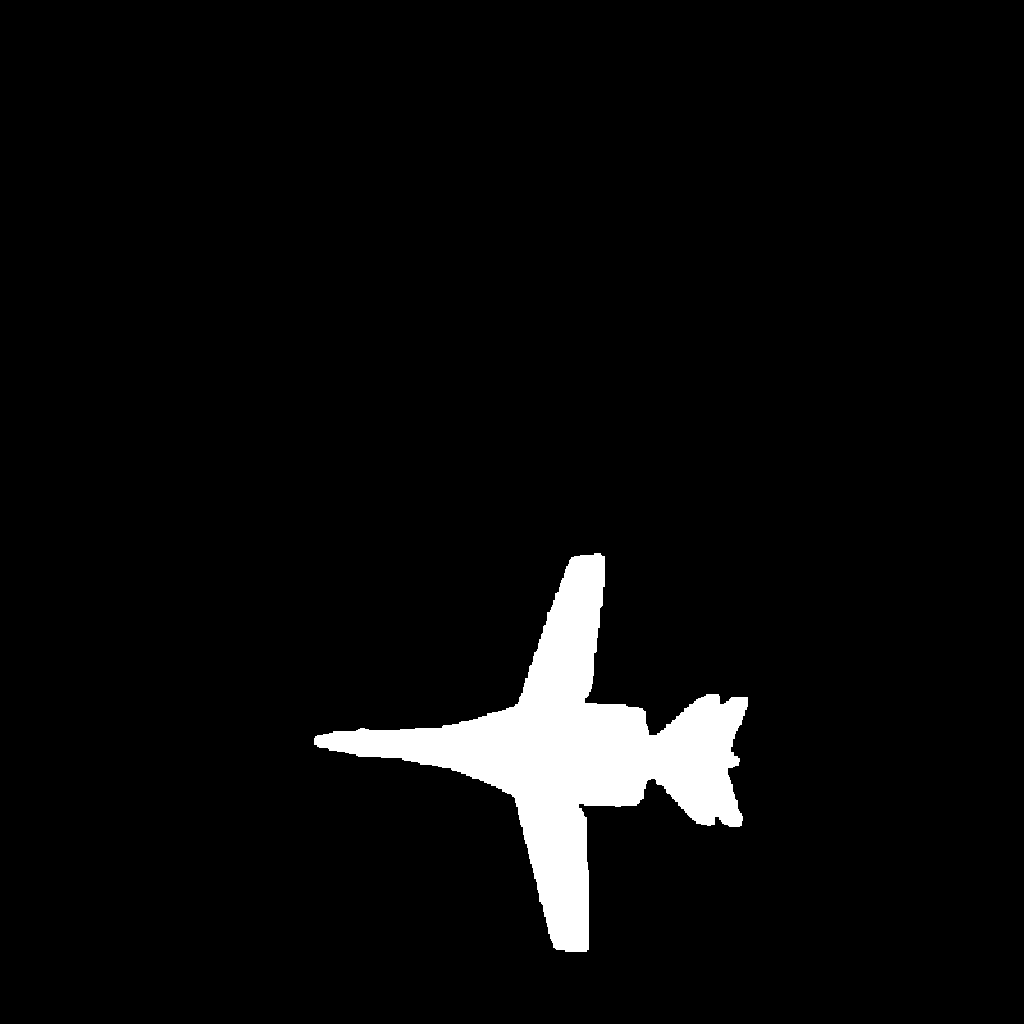}
\end{tabular}
\vspace{-9pt}
\begin{tcolorbox}[colback=gray!20, colframe=black!0, top=0.4pt, bottom=0.4pt, width=0.986\linewidth]
\scriptsize Prompt: \textit{A airplane at the bottom.}
\end{tcolorbox}
\vspace{-3pt}
\begin{tabular}{cccc}
\includegraphics[width=0.232\linewidth]{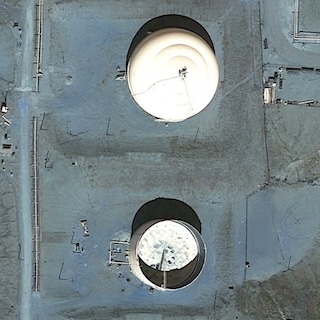} &
\includegraphics[width=0.232\linewidth]{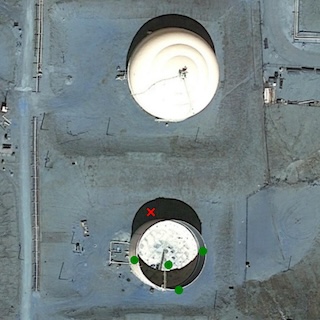} &
\includegraphics[width=0.232\linewidth]{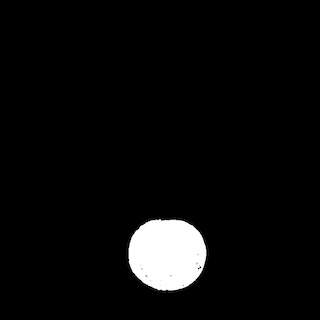} &
\includegraphics[width=0.232\linewidth]{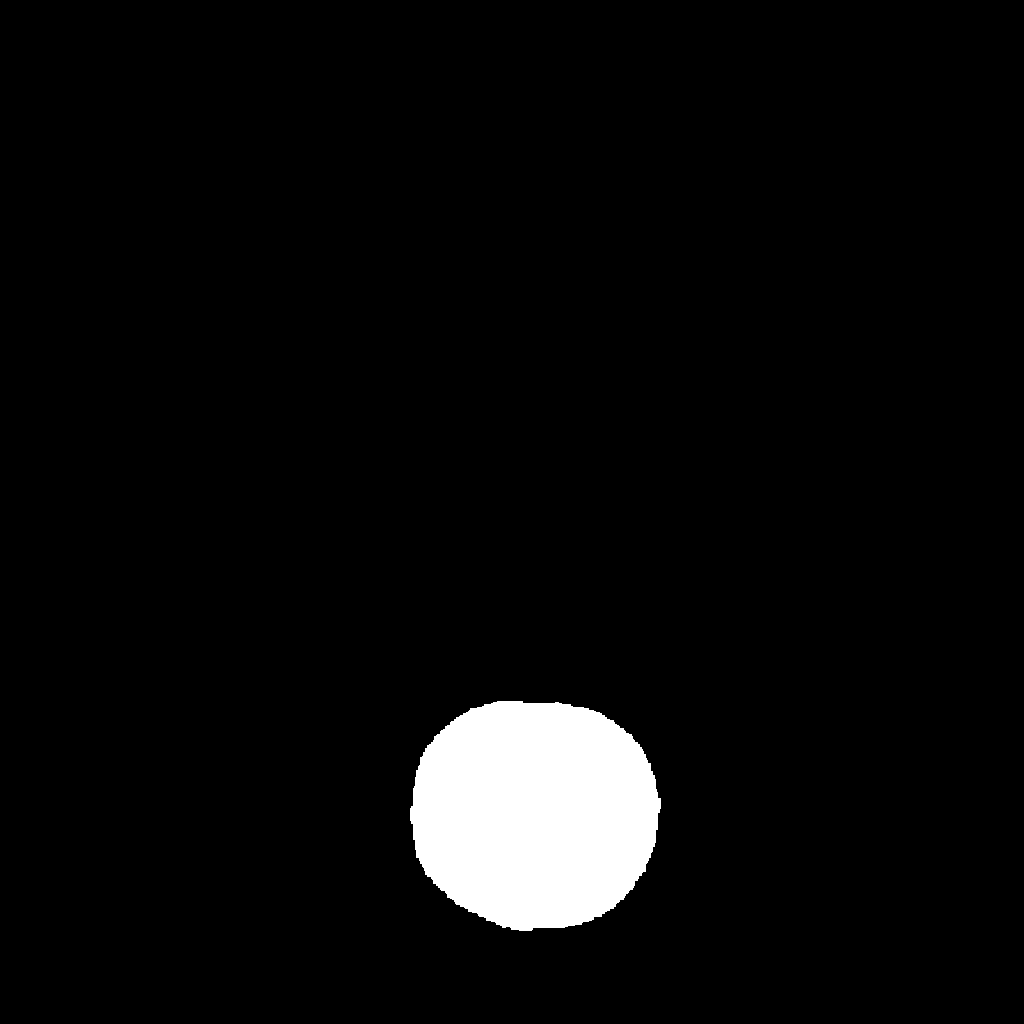}
\end{tabular}
\vspace{-9pt}
\begin{tcolorbox}[colback=gray!20, colframe=black!0, top=0.4pt, bottom=0.4pt, width=0.986\linewidth]
\scriptsize Prompt: \textit{A storage tank at the bottom.}
\end{tcolorbox}
\vspace{-3pt}
\begin{tabular}{cccc}
\includegraphics[width=0.232\linewidth]{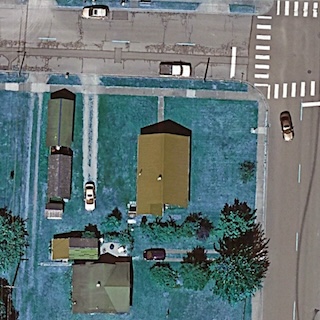} &
\includegraphics[width=0.232\linewidth]{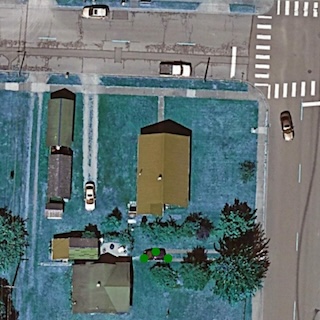} &
\includegraphics[width=0.232\linewidth]{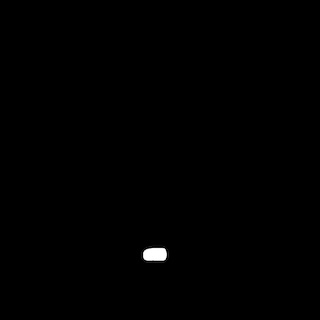} &
\includegraphics[width=0.232\linewidth]{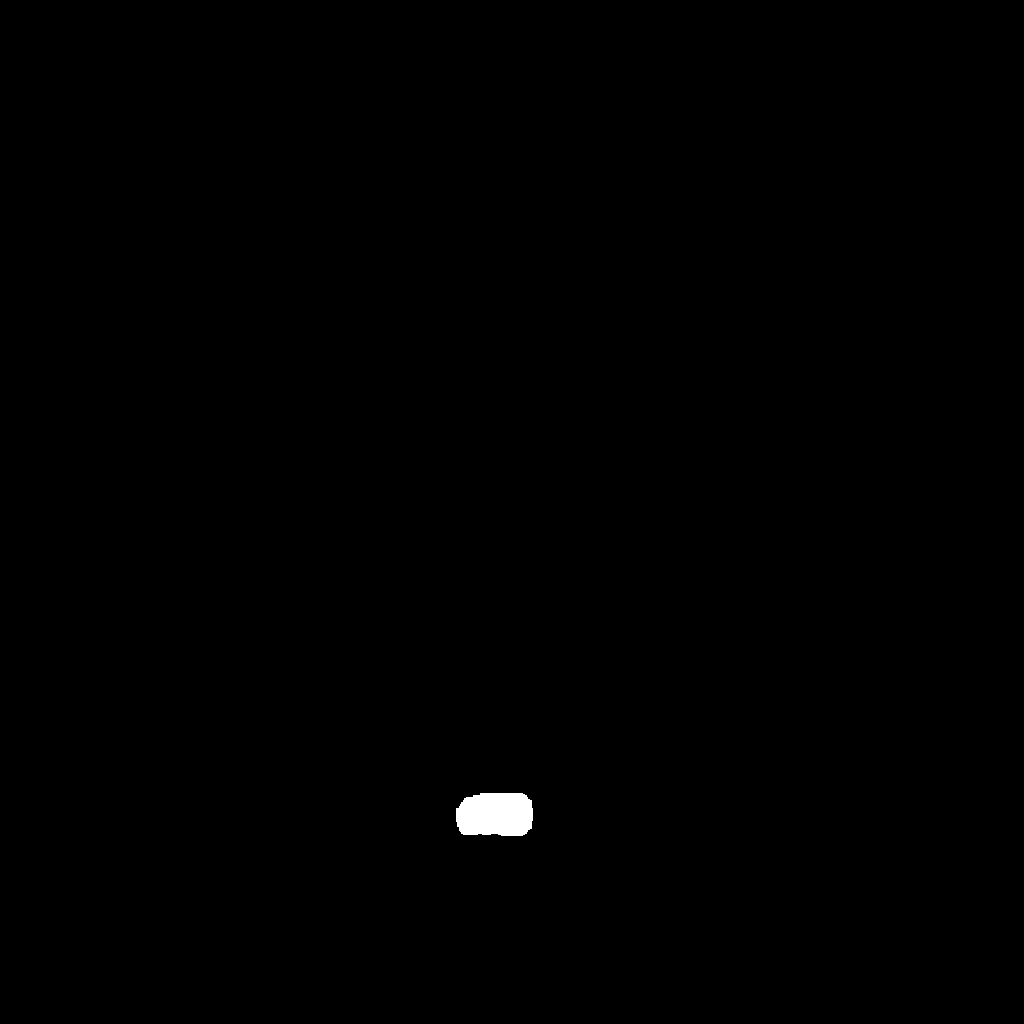}
\end{tabular}
\vspace{-9pt}
\begin{tcolorbox}[colback=gray!20, colframe=black!0, top=0.4pt, bottom=0.4pt, width=0.986\linewidth]
\scriptsize Prompt: \textit{The vehicle at the bottom.}
\end{tcolorbox}
\vspace{-3pt}
\caption{Qualitative results for referring segmentation on RRSIS-D dataset.}
\label{fig:sup-referring-qualitative}
\end{figure}

\autoref{fig:sup-reasoning-comparison} shows additional visualisations of our generative VLM-based approach for reasoning-based segmentation. We visually compare our predicted masks with those produced by the recent SegEarth-R1~\cite{li2025segearth-r1} and GPT-Image-1~\cite{openai_image_generation_2025}. In addition, ~\autoref{fig:sup-reasoning-qualitative} shows more qualitative results for reasoning-based segmentation on EarthReason, while ~\autoref{fig:sup-referring-qualitative} depicts visualisations for referring segmentation using RRSIS-D dataset.

\begin{figure}
\centering \footnotesize \setlength{\tabcolsep}{2pt}
\begin{tabular}{cccc}
Input image & Predicted clicks & Predicted mask & Ground truth \\
\includegraphics[width=0.232\linewidth]{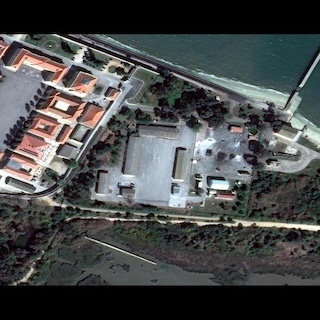} &
\includegraphics[width=0.232\linewidth]{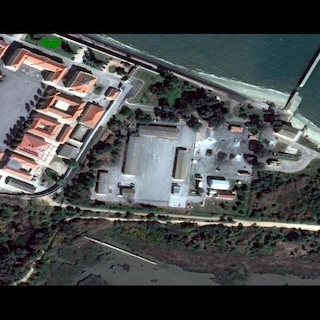} &
\includegraphics[width=0.232\linewidth]{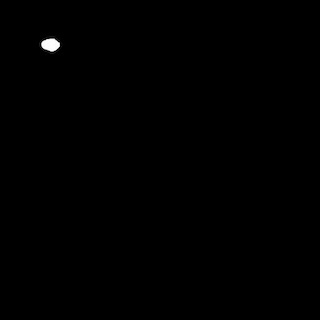} &
\includegraphics[width=0.232\linewidth]{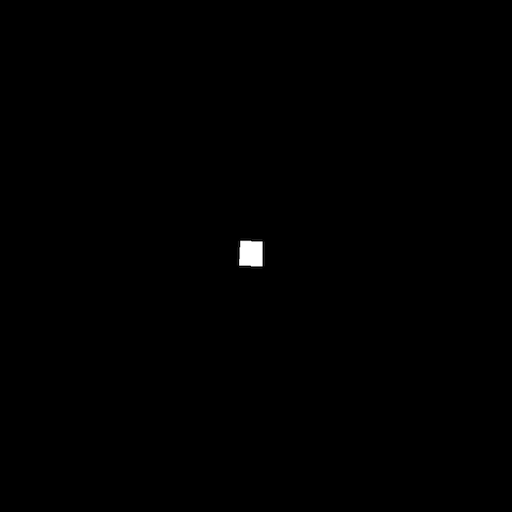}
\end{tabular}
\vspace{-9pt}
\begin{tcolorbox}[colback=gray!20, colframe=black!0, top=0.4pt, bottom=0.4pt, width=0.986\linewidth]
\scriptsize Prompt: \textit{In a situation where a rescue mission is launched, what infrastructure is fundamental for bolstering the capabilities of air units?}
\end{tcolorbox}
\vspace{-3pt}
\begin{tabular}{cccc}
\includegraphics[width=0.232\linewidth]{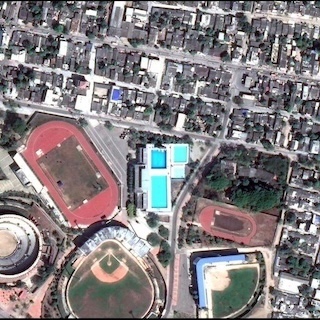} &
\includegraphics[width=0.232\linewidth]{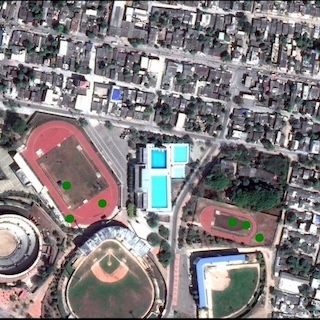} &
\includegraphics[width=0.232\linewidth]{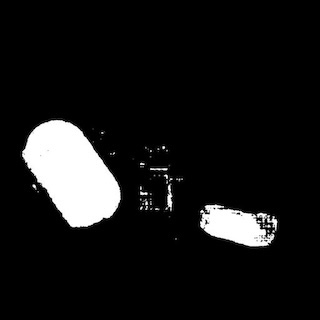} &
\includegraphics[width=0.232\linewidth]{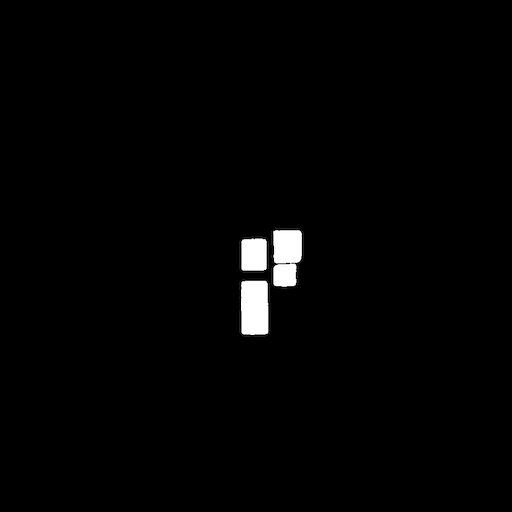}
\end{tabular}
\vspace{-9pt}
\begin{tcolorbox}[colback=gray!20, colframe=black!0, top=0.4pt, bottom=0.4pt, width=0.986\linewidth]
\scriptsize Prompt: \textit{In this city design, which facility serves as the primary venue for recreational activities and competitive training, promoting community health and sports growth?}
\end{tcolorbox}
\vspace{-3pt}
\begin{tabular}{cccc}
\includegraphics[width=0.232\linewidth]{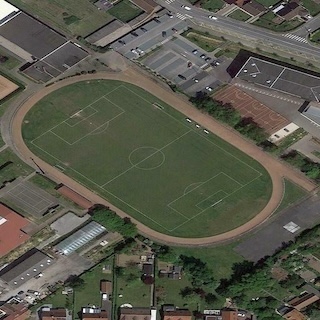} &
\includegraphics[width=0.232\linewidth]{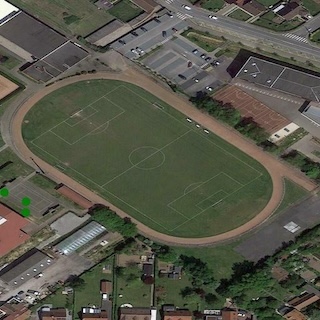} &
\includegraphics[width=0.232\linewidth]{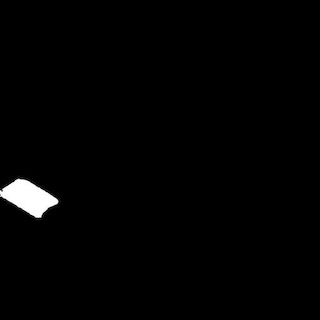} &
\includegraphics[width=0.232\linewidth]{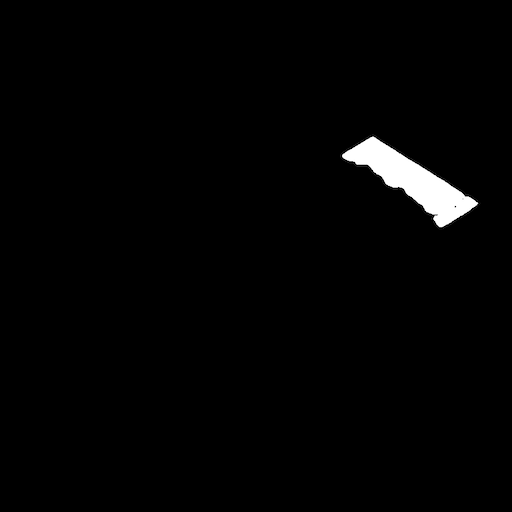}
\end{tabular}
\vspace{-9pt}
\begin{tcolorbox}[colback=gray!20, colframe=black!0, top=0.4pt, bottom=0.4pt, width=0.986\linewidth]
\scriptsize Prompt: \textit{What nearby facility close to the primary recreational zone would be ideal for hosting a casual team sport match on outdoor courts?}
\end{tcolorbox}
\vspace{-3pt}
\begin{tabular}{cccc}
\includegraphics[width=0.232\linewidth]{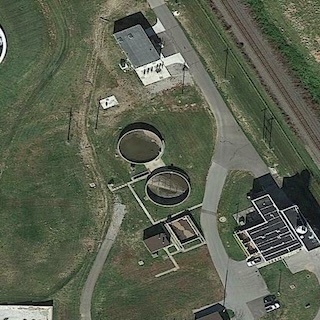} &
\includegraphics[width=0.232\linewidth]{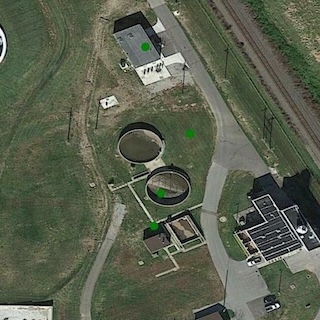} &
\includegraphics[width=0.232\linewidth]{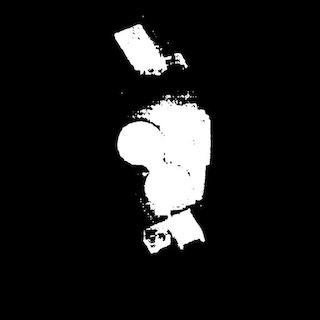} &
\includegraphics[width=0.232\linewidth]{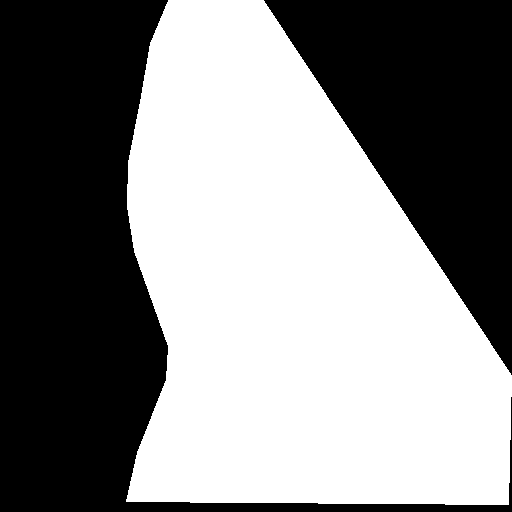}
\end{tabular}
\vspace{-9pt}
\begin{tcolorbox}[colback=gray!20, colframe=black!0, top=0.4pt, bottom=0.4pt, width=0.986\linewidth]
\scriptsize Prompt: \textit{What locations are available for accessing the services that address water pollution in neighborhoods?}
\end{tcolorbox}
\vspace{-3pt}
\begin{tabular}{cccc}
\includegraphics[width=0.232\linewidth]{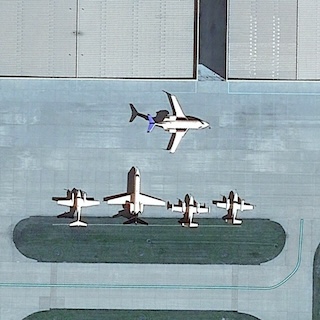} &
\includegraphics[width=0.232\linewidth]{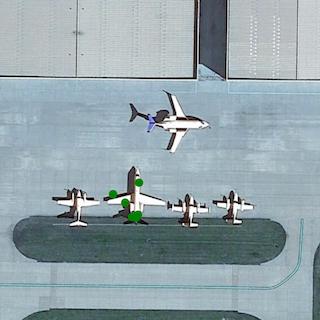} &
\includegraphics[width=0.232\linewidth]{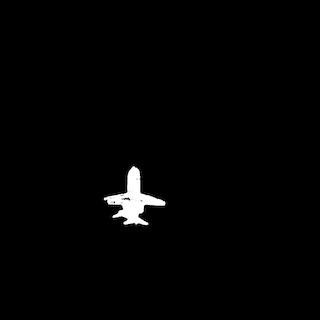} &
\includegraphics[width=0.232\linewidth]{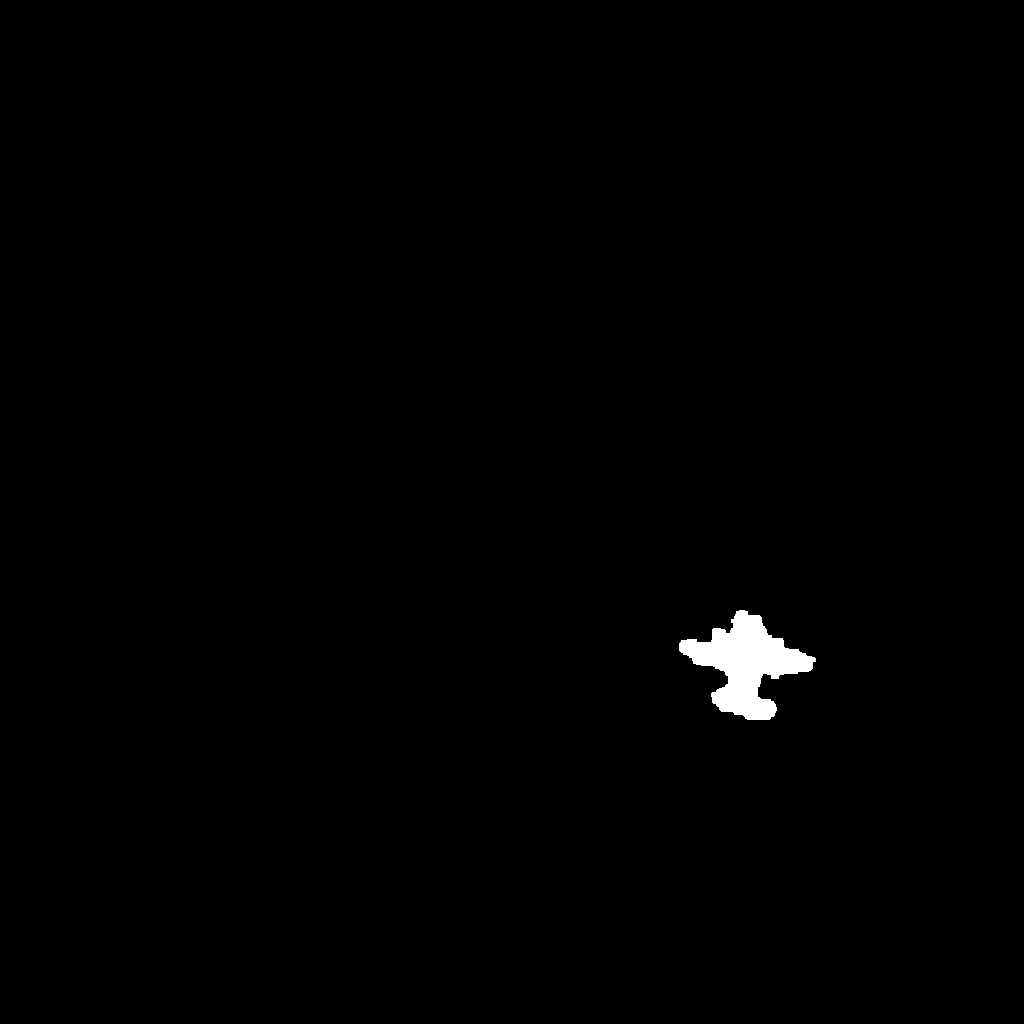}
\end{tabular}
\vspace{-9pt}
\begin{tcolorbox}[colback=gray!20, colframe=black!0, top=0.4pt, bottom=0.4pt, width=0.986\linewidth]
\scriptsize Prompt: \textit{A white airplane.}
\end{tcolorbox}
\vspace{-3pt}
\begin{tabular}{cccc}
\includegraphics[width=0.232\linewidth]{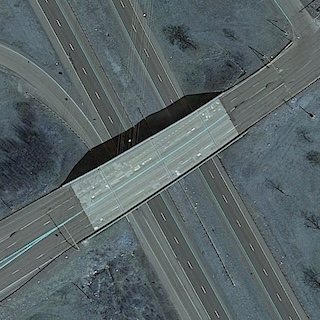} &
\includegraphics[width=0.232\linewidth]{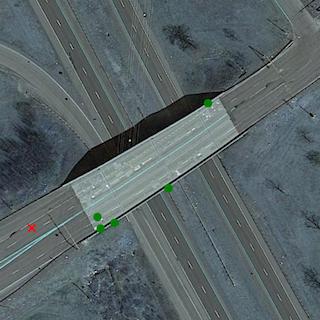} &
\includegraphics[width=0.232\linewidth]{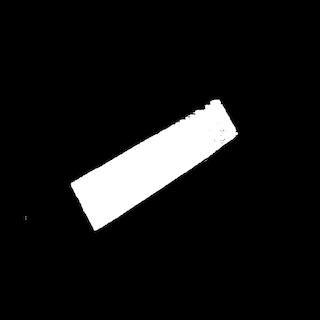} &
\includegraphics[width=0.232\linewidth]{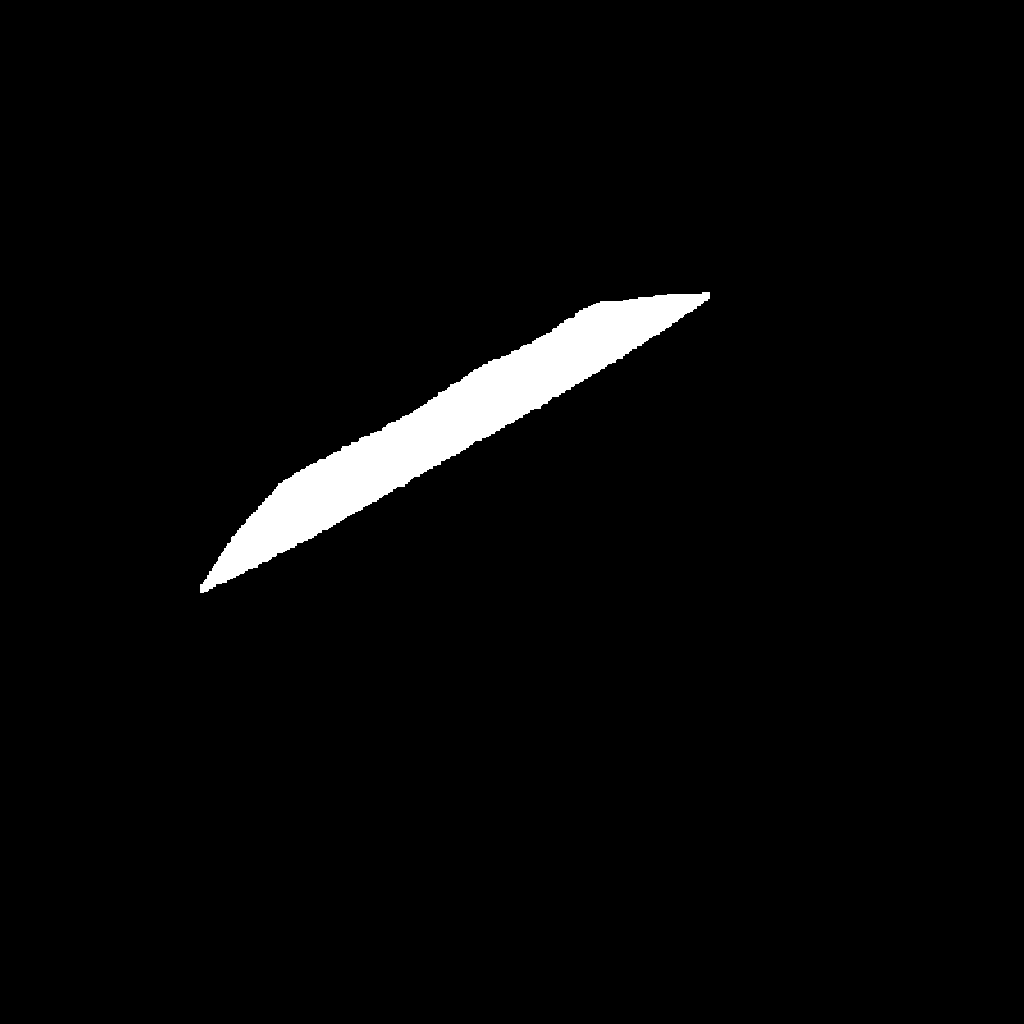}
\end{tabular}
\vspace{-9pt}
\begin{tcolorbox}[colback=gray!20, colframe=black!0, top=0.4pt, bottom=0.4pt, width=0.986\linewidth]
\scriptsize Prompt: \textit{The large overpass in the middle.}
\end{tcolorbox}
\vspace{-3pt}
\caption{Failure cases on the EarthReason and RRSIS-D datasets for reasoning and referring segmentation, respectively.}
\label{fig:sup-failure}
\end{figure}

\paragraph{Failure cases.} \autoref{fig:sup-failure} highlights difficult scenarios. In these cases, the generative VLM-based model produces clicks that lead to segmentation masks with low IoU in both referring and reasoning tasks. We observe three recurring failure modes in the reasoning examples. First, the model might produce regions that are plausible but not the correct answer according to the question, as seen in rows one and three. Second, some questions require additional contextual understanding, as in row two. Here, the model selects an area that is semantically reasonable for answering the question (venues for recreational activities). However, the ground-truth annotation indicates a different region. Finally, we observe cases in which the model selects suitable click locations as in row 4. The target region, however, involves multiple, poorly delimited areas, preventing the segmentation model (SAM) from producing a consistent mask. In referring segmentation scenarios, some errors arise from ambiguous descriptions or annotations, such as the example shown in row 5.

\section{Datasets}
\label{sec:sup-datasets}

\begin{table*}
\centering
\resizebox{\textwidth}{!}{
\begin{tabular}{llllcc}
\toprule
& Dataset & Source / Sensor & Resolution & Classes (FG+BG) & Split Used \\
\midrule
\multicolumn{5}{l}{\textit{Multi-class Open-Vocabulary Semantic Segmentation}} \\
& OpenEarthMap~\cite{xia2023openearthmap} & Satellite \& aerial imagery (global) & 0.25–0.5 m & 8 + 1 & Val \\
& LoveDA~\cite{wang2021loveda} & Google Earth (urban/rural) & 0.3 m & 6 + 1 & Val \\
& iSAID~\cite{waqas2019isaid}& Google Earth, JL-1, GF-2 & - & 15 + 1 & Val \\
& Potsdam\cite{ISPRS_UrbanSemLab} & Aerial (urban) & 5 cm & 5 + 1 & Val \\
& Vaihingen\cite{ISPRS_UrbanSemLab} & Aerial (urban) & 9 cm & 5 + 1 & Val \\
& UAVid~\cite{lyu2020uavid} & UAV video (4K, slanted view) & - & 5 + 1 & Test \\
& UDD5~\cite{chen2018large} & DJI Phantom 4 UAV & 60–100 m altitude & 4 + 1 & Val \\
& VDD~\cite{cai2025vdd} & DJI Mavic Air II UAV & 50–120 m altitude & 6 + 1 & Test \\
\multicolumn{5}{l}{\textit{Single-class Open-Vocabulary Semantic Segmentation}} \\
\rowcolor{brown!6} & WHUAerial & Aerial imagery (Christchurch, NZ) & 0.075 m & 1 + 1 & Val \\
\rowcolor{brown!6} & WHUSat.II & Satellite imagery (East Asia) & 0.45 m &1 + 1 & Test\\
\rowcolor{brown!6} & Inria & Aerial imagery (global cities) & 0.3 m & 1 + 1 & Val \\ 
\rowcolor{brown!6} & xBD & Satellite imagery (multi-disaster) & 0.8 m & 1 + 1 & Test \\
\rowcolor{gray!10} & CHN6-CUG & Google Earth (Chinese cities) & 0.5 m &1 + 1 & Test \\
\rowcolor{gray!10} & DeepGlobe & Satellite imagery (Asia) & 5 m & 1 + 1 & Val \\
\rowcolor{gray!10} & Massachusetts &  Satellite imagery (urban/suburban/rural) & 1 m & 1 + 1 & Test \\
\rowcolor{gray!10} & SpaceNet & Satellite imagery (Las Vegas, Paris, Shanghai, Khartoum) & 0.3 m & 1 + 1 & Test \\
\rowcolor{blue!6} & WBS-SI & Satellite imagery (global) & - & 1 + 1 & Custom~\cite{li2025segearth} \\
\multicolumn{5}{l}{\textit{Referring Segmentation}} \\
& EarthReason & - & - &  - & Val \& Test \\
\multicolumn{5}{l}{\textit{Reasoning Segmentation}} \\
& RRSIS-D & - & - & - & Val \& Test \\
\bottomrule
\end{tabular}
}
\caption{Summary of datasets used in our paper. Rows highlighted in brown correspond to building extraction datasets, green to road extraction, and blue to flood detection datasets. \textit{FG} and \textit{BG} are for foreground and background classes respectively.}
\label{tab:datasets}
\end{table*}

\begin{table*}
\centering
\resizebox{\textwidth}{!}{
\begin{tabular}{ll}
\toprule
Dataset & Class names \\
\midrule
OpenEarthMap~\cite{xia2023openearthmap} &  Bareland, Rangeland, Developed Space, Road, Tree, Water, Agriculture Land, and Building \\
LoveDA~\cite{wang2021loveda} & Building, Road, Water, Barren, Forest, and Agriculture  \\
iSAID~\cite{waqas2019isaid}& Plane, Ship, Storage Tank, Baseball Diamond, Tennis Court, Basketball Court, Ground Track Field, \\
& Harbor, Bridge, Large Vehicle, Small Vehicle, Helicopter, Roundabout, Soccer Ball Field and Swimming Pool\\
Potsdam~\cite{ISPRS_UrbanSemLab} &  Impervious Surfaces, Buildings, Low Vegetation, Tree, and Car \\
Vaihingen~\cite{ISPRS_UrbanSemLab} &  Impervious Surfaces, Buildings, Low Vegetation, Tree, and Car\\
UAVid~\cite{lyu2020uavid} & Background, Building, Road, Car, Tree, Vegetation, and Human \\
UDD5~\cite{chen2018large} & Vegetation, Building, Road, Vehicle, and Background \\
VDD~\cite{cai2025vdd} & Background, Facade, Road, Vegetation, Vehicle, Roof, and Water \\
WHUAerial~\cite{ji2018fully} & Background, and Building \\
WHUSat.II~\cite{ji2018fully} & Background, and Building \\
Inria~\cite{maggiori2017can} & Background, and Building \\ 
xBD~\cite{gupta2019xbd} & Background, and Building \\
CHN6-CUG~\cite{zhu2021global} & Background, and Road \\
DeepGlobe~\cite{DeepGlobe2018} & Background, and Road \\
Massachusetts~\cite{mnih2013machine} & Background, and Road \\
SpaceNet~\cite{van2018spacenet} & Background, and Road \\
WBS-SI~\cite{Kaggle_WaterBodySegmentation} & Background, and Water \\
\bottomrule
\end{tabular}
}
\caption{Class names of OVSS datasets.}
\label{tab:datasets-ovss-classes}
\end{table*}

\noindent\textbf{OpenEarthMap~\cite{xia2023openearthmap}} provides globally distributed satellite and aerial imagery with a spatial resolution ranging from $0.25$ to $0.5 m$. It comprises 9 classes including background class. We follow~\cite{li2025segearth} setup and evaluate on its validation set, excluding xBD data.

\noindent\textbf{LoveDA~\cite{wang2021loveda}} contains $0.3 m$ resolution images sourced from Google Earth, covering both urban and rural scenes. It includes 6 foreground categories and 1 background class. We use the validation set for evaluation.

\noindent\textbf{iSAID~\cite{waqas2019isaid}} contains $655,451$ annotated object instances across 15 categories in $2,806$ high-resolution images. The dataset features large scale variation, dense object distributions, and imbalanced category frequencies, reflecting real-world aerial conditions. All images are identical to those in DOTA-v1.0~\cite{Xia_2018_CVPR}, primarily collected from Google Earth, JL-1, and GF-2 satellites. We evaluate on its validation set, which consist on $11,644$ image patches.

\noindent\textbf{Potsdam} and \textbf{Vaihingen}~\cite{ISPRS_UrbanSemLab} datasets are designed for urban semantic segmentation appearing in the 2D Semantic Labeling Contest. Their spatial resolutions are $5 cm$ and $9 cm$, respectively, each containing 6 classes. We use their validation sets for evaluation.

\noindent\textbf{UAVid~\cite{lyu2020uavid}} consists of 30 video sequences captured in 4K resolution from oblique urban views. We treat individual frames as independent images and follow merging process of some categories as in~\cite{li2025segearth}. The final dataset contains 5 foreground classes and 1 background class. We evaluate it using its test set.

\noindent\textbf{UDD5~\cite{chen2018large}} is captured by a DJI Phantom 4 UAV flying at altitudes varying from $60$ to $100 m$. It contains 4 foreground categories and 1 background class. We use its validation set for evaluation.

\noindent\textbf{VDD~\cite{cai2025vdd}} is collected with a DJI Mavic Air II drone, comprising 400 RGB images with a resolution of $4000 \times 3000$ pixels. The images are taken from altitudes between $50 m$ and $120 m$. It includes 6 foreground classes and 1 background class. We evaluate on its test set.

\noindent\textbf{WHUAerial~\cite{ji2018fully}} comprises manually curated aerial and satellite imagery collections for building extraction. The aerial subset contains over 220 k individual building instances extracted from high-resolution imagery $(0.075 m)$ covering approximately $450 km^2$ of Christchurch, New Zealand. We use its validation set for evaluation.

\noindent\textbf{WHUSat.II~\cite{ji2018fully}} includes six adjacent satellite images covering $860 km^2$ across East Asia with $0.45$ m ground resolution. It contains $34,085$ manually annotated buildings, cropped into $17,388$ non-overlapping tiles. This subset is specifically designed to assess model generalisation across similar building styles from different data sources within the same geographic region. We use its test set for evaluation.

\noindent\textbf{Inria~\cite{maggiori2017can}} dataset contains diverse urban environments, from dense city centers (\eg, San Francisco) to alpine towns (\eg, Lienz). Each subset contains distinct cities, enabling evaluation of cross-region generalisation under varying geographic, illumination, and seasonal conditions. It covers 810 $km^2$ with a spatial resolution of $0.3m$. We use the test set for evaluation.

\noindent \textbf{xBD~\cite{gupta2019xbd}} is a large-scale, high-resolution satellite imagery benchmark designed for building damage assessment and change detection in disaster scenarios. It comprises $22,068$ pre and post-disaster image pairs covering 19 natural hazard events, spanning a total area of $45,362 km^2$ and including $850,736$ annotated building footprints. Its spatial resolution is $0.8m$. We use the pre-disaster satellite data of test set for evaluation.

\noindent\textbf{CHN6-CUG~\cite{zhu2021global}} is a large-scale, manually annotated benchmark for pixel-level road extraction from satellite imagery, collected from Google Earth. It includes imagery from six representative Chinese cities capturing diverse levels of urbanisation and road structures. It contains 4511 labeled images with a spatial resolution of 0.5m. We use its test set for evaluation.

\noindent\textbf{DeepGlobe~\cite{DeepGlobe2018}} provides high-resolution (0.5 m) satellite imagery sampled from the DigitalGlobe + Vivid collection, covering regions in Thailand, Indonesia, and India. It contains 8,570 RGB images spanning $2,220 km^2$ in total. Pixel-level annotations delineate road and background classes, capturing diverse surfaces and urban–rural variations. We use the validation set for evaluation.

\noindent\textbf{Massachusetts~\cite{mnih2013machine}} is an aerial imagery dataset for road segmentation, designed to address challenges such as occlusions from trees, building shadows, and road texture variations. It contains 1,171 aerial RGB images, covering a total area of $2,600 km^2$. Road labels are generated by rasterising OpenStreetMap centerlines with a 7-pixel width at 1m spatial resolution. The dataset includes diverse urban, suburban, and rural scenes. We use its test set for evaluation.

\noindent\textbf{SpaceNet~\cite{van2018spacenet}} contains satellite imagery with a spatial resolution of 0.3m, covering Las Vegas, Paris, Shanghai, and Khartoum. It was introduced for the SpaceNet Road Detection and Routing Challenge, designed to support automated road extraction from very high-resolution imagery. We use the test set for evaluation.

\noindent\textbf{WBS-SI~\cite{Kaggle_WaterBodySegmentation}} is a satellite imagery dataset designed for water body segmentation. Following the setup in~\cite{li2025segearth}, we perform our evaluation using their proposed test split.

\noindent\textbf{EarthReason}~\cite{li2025segearth-r1} is the first large-scale benchmark dataset designed for geospatial pixel reasoning. It contains 5,434 high-resolution remote sensing images, each annotated with a manually created segmentation mask targeting specific regions of interest. Alongside these masks, the dataset includes over 30,000 implicit question-answer pairs that require spatial understanding and reasoning to identify the correct target regions. The dataset is divided into training, validation, and test splits of 2,371, 1,135, and 1,928 images, respectively.

\noindent\textbf{RRSIS-D}~\cite{liu2024rotated} is a dataset designed for Referring Remote Sensing Image Segmentation (RRSIS), specifically to handle significant variations in spatial resolution and object orientation. The dataset is constructed by converting bounding box annotations from the RSVGD dataset~\cite{zhan2023rsvg} into instance masks. It comprises 20 semantic categories, including aircraft, golf courses, highway service areas, baseball fields, and stadiums. It is further extended with 7 descriptive attributes to enhance the clarity and expressiveness of referring expressions. RRSIS-D exhibits scale variability, with some targets covering minimal pixel areas and others exceeding 400,000 pixels. Overall, the dataset comprises 17,402 image-description-mask triplets, divided into 12,181 for training, 1,740 for validation, and 3,481 for testing.

{
    \small
    \bibliographystyle{ieeenat_fullname}
    \bibliography{main}
}

\end{document}